\documentclass[journal,twoside,web]{ieeecolor}
\usepackage{generic}
\usepackage{hyperref}
\usepackage{textcomp}
\def\BibTeX{{\rm B\kern-.05em{\sc i\kern-.025em b}\kern-.08em
    T\kern-.1667em\lower.7ex\hbox{E}\kern-.125emX}}
\usepackage{amssymb}                                                       
\usepackage{ifpdf}
\usepackage{amsmath}
\usepackage{amsfonts}
\usepackage{mathrsfs}
\usepackage{graphicx}
\usepackage{subfigure}
\usepackage{epstopdf}
\usepackage{lineno}
\usepackage{cite}
\usepackage{enumerate}
\usepackage{algorithm, algorithmic}
\usepackage{bm}
\usepackage{tikz}
 \usepackage{xcolor}

\newtheorem{theorem}{Theorem}
\newtheorem{definition}{Definition}

\newtheorem{lemma}{Lemma}     
\newtheorem{remark}{Remark}

\newtheorem{assumption}{Assumption}

\newcommand{\brk}[1]{\left[#1\right]}

\begin{document}
\title{Risk-Averse Learning with Delayed Feedback}
\author{Siyi Wang, Zifan Wang, Sandra Hirche, \IEEEmembership{Fellow, IEEE}, and Karl H. Johansson, \IEEEmembership{Fellow, IEEE}
\thanks{This work was supported  by the European Research Council
(ERC) Consolidator Grant ”Safe data-driven control for humancentric systems (CO-MAN)” under grant agreement number 864686, by the Swedish Research Council Distinguished Professor Grant 2017-01078, Knut and Alice Wallenberg Foundation, Wallenberg Scholar Grant, the Swedish Strategic Research Foundation SUCCESS, AFOSR under award \#FA9550-19-1-0169, and  NSF under award CNS-1932011.}
\thanks{Siyi~Wang, Zifan~Wang, and Karl H. Johansson are with the Division of Decision and Control Systems, School of Electrical Engineering and Computer Science, KTH Royal Institute of Technology,
10044  Stockholm, Sweden, e-mail: \{siyiw,zifanw, kallej\}@kth.se.}
\thanks{Sandra~Hirche is with the Chair of Information-oriented Control (ITR), Technical University of Munich, Germany, e-mail: \{hirche\}@tum.de.}}

\maketitle

\begin{abstract}
In real-world scenarios, risk-averse learning is valuable for mitigating potential adverse outcomes. However,  delayed feedback makes it challenging to assess and manage risk effectively.
In this paper, we investigate risk-averse learning using Conditional Value at Risk (CVaR) as risk measure, while incorporating feedback with random yet bounded delays.   
We develop two risk-averse learning algorithms that rely on one- and two-point zeroth-order optimization, respectively.  
The dynamic regrets of the algorithms are analyzed in terms of the cumulative delay and the total number of samples.  
In the absence of delay, 
the regret bounds match the established bounds of zeroth-order stochastic gradient methods for risk-averse learning. 
When the maximum delay is independent of the horizon, 
the two-point regret bound is no worse than the one-point bound and can attain a strictly better order 
with a sufficiently large sampling budget.
We provide numerical experiments on a dynamic pricing problem to demonstrate the performance of the algorithms.
\end{abstract}


\section{Introduction}\label{sec:introduction}
In online learning, the decision-maker sequentially updates its decision in a changing environment relying on historical information such as observations of previous actions and costs \cite{hazan2016introduction}.
However, in real-world scenarios, these observations may arrive with delays and impact system performance.
The performance of online learning is typically measured through regret, 
which quantifies the performance gap between the algorithm's decisions and a chosen benchmark.
Static regret \cite{flaxman2004online} compares against the best fixed decision in hindsight, 
whereas dynamic regret compares the algorithm's performance against a time-varying sequence of optimal decisions, 
providing a more informative measure in non-stationary or evolving settings \cite{zinkevich2003online,yi2022regret}. 
Common metrics used to analyze dynamic regret include function variation \cite{besbes2015non}
and the path length of the optimal decisions \cite{zinkevich2003online}.

Unlike traditional learning that optimizes expected outcomes, 
risk-averse learning focuses on extreme outcomes to mitigate catastrophic outcomes under uncertainties \cite{cardoso2019risk,greenberg2022efficient,ahmadi2023risk}. 
Its application includes financial investment \cite{alexander2006minimizing}, power grid management \cite{tavakoli2018cvar}, and robotics \cite{ahmadi2021risk}.
Among various risk measures, conditional value at risk (CVaR) is widely adopted, as it offers a coherent risk assessment by considering the tail end of the loss distribution \cite{rockafellar2000optimization,chapman2021risk}. 
In general, the CVaR objective and its gradient are not directly observable. 
The learner has access only to stochastic loss evaluations at queried actions.
Zeroth-order methods therefore provide a natural way to estimate the CVaR gradient from these function evaluations \cite{flaxman2004online,shames2019online}.
The pioneering work \cite{flaxman2004online} applies zeroth-order algorithms to online convex optimization. 
This method evaluates the function at a single point per iteration,  and is known as one-point zeroth-order optimization. 
Based on this, various approaches have been developed to enhance the estimation performance, 
including the one-point residual \cite{zhang2022new} and multi-point zeroth-order optimization \cite{shamir2017optimal,nesterov2017random}. 

Feedback in online systems may arrive late because of sensing, aggregation, communication, or computation latency \cite{fridman2014introduction,liu2019survey}. 
For instance, in online learning platforms, the impact of an educational intervention might be observed with a time lag. 
In medical diagnosis, the efficacy of a treatment plan might become apparent after several weeks or months. 
In online recommender systems, click-through rates are aggregated and reported back at regular intervals.
Accounting for delays is essential, as relying on outdated information degrades system performance, 
especially in risk-averse settings where timely data is essential for making decisions.

A substantial literature therefore studies online learning under various delay models 
\cite{langford2009slow,li2019bandit,heliou2020gradient,vernade2020non,joulani2013online,howson2023delayed,wan2023non,wang2021delay,wang2022delay}. 
To name a few, \cite{li2019bandit} studies bandit methods with random delay,
and achieves a static regret bound of
$\widetilde{\mathcal O}(\sqrt{K(T+D)})$, where $K$ denotes the
number of arms or the decision dimension, and $D$ is the cumulative
delay.
The paper \cite{heliou2020gradient} studies bandit online learning in multi-agent games,
and show that the algorithm converges to Nash equilibria when the delays are tame.
In particular, this reduces to
$\widetilde{\mathcal O}(T^{3/4})$ under bounded delays.
Some work, e.g., \cite{joulani2013online,howson2023delayed}, investigate multi-armed bandit learning with delayed feedback, where the delays are independent and identically distributed (i.i.d.) random variables with a finite expectation \cite{joulani2013online} and subexponential \cite{howson2023delayed}, respectively.
Notably, the above works evaluate algorithm performance using static regret.
However, delayed feedback poses additional challenges in changing environments, 
making its analysis under dynamic settings particularly important.
Only few papers, such as \cite{wang2021delay,wang2022delay,wan2023non},
have studied online learning using the dynamic regret,
addressing fixed constant delay \cite{wang2021delay,wang2022delay} and arbitrary delay \cite{wan2023non}. 
For deterministic first-order online convex optimization with arbitrary delays, 
\cite{wan2023non} establishes a dynamic regret bound of $\mathcal O(\sqrt{d_{\mathrm{avg}}T(P_T+1)})$ 
under mild conditions and the worst-case optimal bound $\mathcal O(\sqrt{d_{\max}T(P_T+1)})$, 
where $d_{\mathrm{avg}}$ and $d_{\max}$ are the average and maximum delays, respectively, and $P_T$ is the comparator path length.
Additionally, \cite{wang2021delay,wang2022delay} study deterministic constrained optimization with time-varying objective and a fixed delay $\tau$.
These results are not directly comparable with ours under identical
assumptions. Most work consider
risk-neutral static regret, whereas
\cite{wan2023non,wang2021delay,wang2022delay} assume deterministic
first-order feedback. In contrast, we consider dynamic regret under a
stochastic CVaR objective with delayed zeroth-order feedback, 
where each delayed packet contains multiple stochastic samples.

This paper studies risk-averse online learning when stochastic zeroth-order feedback is subject to time-varying yet bounded delays. 
At each iteration, the learner queries a batch of i.i.d. stochastic cost samples at each perturbed action. 
Once the delayed feedback arrives, 
these samples are used to construct empirical distributions and the corresponding empirical CVaR estimates, 
from which the one- and two-point zeroth-order gradient estimates are formed.
The delay may cause packets to arrive out of order, multiple packets may arrive simultaneously, 
and some packets may remain unavailable at the end of the horizon. 
We therefore organize the received packets on a virtual timeline according to their utilization order and define the corresponding cumulative delay. 
Since the empirical CVaR errors depends on the sample batch size and accumulates over the delayed virtual updates, 
we index the filtrations by this virtual order to establish the conditional independence and concentration arguments. 
Based on this, we develop one- and two-point zeroth-order algorithms and establish high-probability dynamic-regret bounds
in terms of the cumulative delay, sampling budget, and comparator path length. 
The bounds improve as the sampling budget increases or the delay decreases. 
Moreover, when the maximum delay is independent of the horizon, the
two-point regret bound is no worse than the one-point bound and can
attain a strictly better order with a large sampling budget.

The remainder of this article is structured as follows: Section~\ref{sec:preliminary} introduces preliminaries and problem formulation. Sections~\ref{sec:one-point} and~\ref{sec:two-point} present the main results on one- and two-point risk-averse learning with delayed feedback, respectively. Section~\ref{sec:simulation} demonstrates the performance of the algorithms through numerical simulations. Section~\ref{sec:conclusion} draws conclusions. \\
\textbf{Notations:} Let $\|\cdot\|$ denote the $l_2$ norm. Let the notation $\mathcal{O}$ hide the constant and $\tilde{\mathcal{O}}$ hide constant and polylogarithmic factors of the number of iterations $T$, respectively.
Let $A \oplus B = \{a + b \vert a \in A, b \in B\}$ denote the Minkowski sum of two sets of position vectors $A$ and $B$ in Euclidean space.


\section{Preliminaries and problem statements}\label{sec:preliminary}
Consider the cost function $J_t(x,\xi):\mathcal X\times\Xi\to\mathbb R$
for $t=1,\dots,T$, where $x\in\mathcal X$ denotes the decision variable,
$\xi\in\Xi$ denotes the random noise, and $T$ is the horizon. The
admissible set $\mathcal X\subseteq\mathbb R^d$ is assumed to be
nonempty, compact, and convex. Let
$\mathbb B^d:=\{x\in\mathbb R^d:\|x\|\leq1\}$ denote the unit ball. We
assume that $r\mathbb B^d\subseteq\mathcal X$ for some $r>0$. 
Denote its diameter by $D_x=\sup_{x,y\in\mathcal X}\|x-y\|$.
\subsection{CVaR} 
We use CVaR as the risk measure. Suppose $J_t(x,\xi)$ has the cumulative
distribution function $F_{t,x}(y)=P(J_t(x,\xi)\le y)$, and write $F_t$
when the action is clear from context. Assume that the loss is bounded
by $U>0$, i.e., $|J_t(x,\xi)|\le U$. Given a risk level
$\alpha\in(0,1]$,
the $\mathrm{CVaR}_{\alpha}$ measures the expected loss in the upper
$\alpha$-tail of $J_t(x,\xi)$. We adopt the
Rockafellar--Uryasev representation \cite{rockafellar2000optimization}:
\begin{align}\label{eq:CVaR def}
 C_t(x)
&:=\mathrm{CVaR}_{\alpha}\left[J_t(x,\xi)\right]\nonumber\\
& :=\inf_{z\in\mathbb R}\left\{
z+\frac{1}{\alpha}\mathbb E\left[(J_t(x,\xi)-z)_+\right]
\right\}.
\end{align} This definition remains valid when the loss distribution has atoms,
and hence also applies to the empirical distributions used below.
Since the exact CVaR gradient is generally unavailable, we construct its estimate using a zeroth-order optimization approach that queries the objective function at perturbed actions \cite{flaxman2004online}. We perturb the action $\hat{x} = x + \delta u$  and query the loss $J_t(\hat{x},\xi)$, where $u$ is sampled uniformly from the unit sphere $\mathbb{S}^{d-1} $ and $\delta$ is the perturbation radius, also known as the smoothing parameter.

\subsection{Delay model}\label{sec:delay} 

Assume that the loss $J_t(\hat{x}_t,\xi_t)$ will be observed after $d_t$
time slots, where $d_t$ is a nonnegative integer that can vary from slot
to slot.
For each time slot $t$, the learner collects the losses and the perturbation direction vectors, 
which is denoted as $\mathcal{L}_t$.  
When the learner does not receive loss at time $t$, we have $\mathcal{L}_t = \emptyset$. 
When the learner receives multiple losses at time $t$, 
we denote $\Delta_t$ as the number of losses received at time $t$. 
At time $t$, denote $s_{n|t}$, for $n=1,\dots,\Delta_t$, as the generating time of the arrival feedback.
We sort the received losses by their generation time. 
Then, the collection of the learner at time $t$ is 
$\mathcal{L}_t = \big\{\big(J_{s_{n|t}}(\hat{x}_{s_{n|t}}), u_{s_{n|t}} \big)\ 
|\ s_{n|t} + d_{s_{n|t}} = t, s_{n|t} \le t, n =1,\dots, \Delta_t \big\}$. At time $t$, the realized delay $d_{s_{n|t}}$ of the arrival feedback is unknown to the learner. 
See Fig.~\ref{fig:delay illustration} for an illustrative example. 

\begin{figure}
    \centering
        \scalebox{0.65}{
    \begin{tikzpicture}[
  timeline/.style={draw, thick}, 
  dashedtimeline/.style={draw, thick, dashed}, 
  event/.style={draw, fill=blue!20, circle, minimum size=2mm, inner sep=0}, 
  label/.style={font=\small, align=center} 
]

\coordinate (start) at (0,0);
\coordinate (end) at (12,0);
\coordinate (top) at (0,2); 
\coordinate (topend) at (12,2); 

\draw[dashedtimeline] (start) -- (1,0);  
\draw[timeline] (1,0) -- (11,0); 
\draw[dashedtimeline] (11,0) -- (end);  

\draw[dashedtimeline] (top) -- (1,2); 
\draw[timeline] (1,2) -- (11,2); 
\draw[dashedtimeline] (11,2) -- (topend); 

\node[below=1.2cm, font=\small] at (0.6,1) {\textsf{Collection}};
\node[event] (event1) at (2,0) {};
\node[label, below=2mm] at (event1) {$\mathcal{L}_{t-2}=\emptyset$};

\node[event] (event2) at (5,0) {};
\node[label, below=2mm] at (event2) {$\mathcal{L}_{t-1}=\emptyset$};

\node[event] (event3) at (8,0) {};
\node[label, below=2mm] at (event3) {$\mathcal{L}_{t}=\{(J_{t-2},u_{t-2}),\}$};
\node[label, below=6mm] at (event3) {$(J_{t},u_{t} )\}$};

\node[event] (event4) at (11,0) {};
\node[label, below=2mm] at (event4) {$\mathcal{L}_{t+1} = $};
\node[label, below=6mm] at (event4) {$\{(J_{t-1},u_{t-1}) \}$};

\node[above=1.2cm, font=\small] at (0.7,1) {};
\node[event] (event5) at (2,2) {};
\node[label, above=2mm] at (event5) {$(J_{t-2},u_{t-2})$};

\node[event] (event6) at (5,2) {};
\node[label, above=2mm] at (event6) {$(J_{t-1},u_{t-1})$};

\node[event] (event7) at (8,2) {};
\node[label, above=2mm] at (event7) {$(J_t,u_t)$};

\node[event] (event8) at (11,2) {};
\node[label, above=2mm] at (event8) {$(J_{t+1},u_{t+1})$};

\draw[dashed] (event1) -- (event5);
\draw[dashed] (event2) -- (event6);
\draw[dashed] (event3) -- (event7);
\draw[dashed] (event4) -- (event8);

\draw[->] (event5) -- (event3);
\draw[->] (event6) -- (event4);
\draw[->] (event7) -- (event3);
\draw[dashed][->] (event8) -- (12,1);
    \end{tikzpicture}}
    \caption{An example illustrates the delay model: the learner does not receive feedback at slot $t-1$, i.e., $\mathcal{L}_{t-1}=\emptyset$, 
    and receives two feedbacks at slot $t$, which are generated at slots $t-2$ and $t$, respectively, 
    i.e., $\mathcal{L}_{t}=\{J_{s_{1|t}},u_{s_{1|t}}, J_{s_{2|t}},u_{s_{2|t}} \}$. with $s_{1|t} = t-2$ and $s_{2|t} = t $.}
    \label{fig:delay illustration}
\end{figure}

\subsection{Problem statement}  
We make the following assumptions.
\begin{assumption}\label{ass:max delay} 
The delay is bounded, i.e., $d_t\le\bar d$ for
$t\in\{1,\dots,T\}$, where the delay bound $\bar d\ge1$ is an integer.
\end{assumption}
Assumption~\ref{ass:max delay} imposes a hard upper bound on the feedback latency. In
practice, this bound may be specified by a system-level setting, and is
assumed to be available before learning. 
\begin{assumption}\label{assumption:convex}
The cost function $J_t(x,\xi) $ is convex in $x$ for every $\xi \in \Xi$. 
\end{assumption}
\begin{assumption}\label{assumption:Lipschitz}
The cost function $J_t(x,\xi)$ is $L_0$-Lipschitz continuous in $x$
for every $\xi\in\Xi$. Moreover, it is uniformly bounded, i.e.,
$|J_t(x,\xi)|\le U$ for all $t$, $x\in\mathcal X$, and $\xi\in\Xi$.
\end{assumption}

We evaluate the performance of the algorithm through dynamic regret, which is defined as the cumulative loss under the performed actions against  best actions in hindsight: 
\begin{equation}\label{eq:dynamic regret definition}
    \mathrm{DR}(T)  = \sum_{t=1}^T C_t(\hat{x}_t)-  \sum_{t=1}^T C_t(x_t^\ast),
\end{equation}
where $\hat{x}_t$ is the action generated by the  algorithm at time slot $t$, and $ x_t^\ast = {\textrm{argmin}}_{x \in \mathcal{X}}  C_t (x)$ is the optimal action at time $t$.  
This paper aims to develop risk-averse learning algorithms that achieve a sublinear regret, i.e., $\lim_{T\rightarrow \infty}
\mathrm{DR}(T)/T = 0 $.

\section{One-point risk-averse learning}\label{sec:one-point}
This section will develop a one-point zeroth-order risk-averse algorithm and analyze its performance. 

Since only function evaluations are available, we use the zeroth-order optimization method to estimate the CVaR gradient for the learning process. 
Let $\mathbb{S}^{d-1} = \{u \in \mathbb{R}^d: \| u\| =1\}$ denote the boundary of $\mathbb{B}^d$.
A smoothed approximation \cite{flaxman2004online} of the CVaR function is constructed as
\begin{equation}\label{eq:smoothed cvar}
 C_t^\delta(x) = \mathbb{E}_{u \sim \mathbb{B}^{d}}[C_t(x+\delta u)]. 
\end{equation} 
This definition is used at points $x$ satisfying
$x+\delta\mathbb B^d\subseteq\mathcal X$.
Then, the gradient of the smoothed function \eqref{eq:smoothed cvar} is given by
\begin{equation}\label{eq:one point gradient estimate}
\nabla C_t^\delta(x) = \mathbb{E}_{u\sim \mathbb{S}^{d-1}}\brk{\frac{d}{\delta}C_t(x+\delta u)u}.
\end{equation}
The following lemmas analyze the CVaR function and its smoothed approximation, which are prepared for the regret analysis. 
\begin{lemma}\label{lemma:cvar convex}\cite{cardoso2019risk}
Given Assumption \ref{assumption:convex}, $C_t(x)$ and $C_t^\delta(x)$ are both convex in $x$.
\end{lemma} 

\begin{lemma}\label{lemma:smoothed cvar Lipschitz}\cite{cardoso2019risk}
Given Assumption \ref{assumption:Lipschitz}, both $C_t(x)$ and
$C_t^\delta(x)$ are $L_0$-Lipschitz continuous in $x$, and
$|C_t^\delta(x)-C_t(x)|\le\delta L_0$.
\end{lemma} 
By the CVaR definition~\eqref{eq:CVaR def}, CVaR is a risk
measure of a stochastic cost, whose distribution is characterized by
its cumulative distribution function. Hence, when the underlying
stochastic cost is clear, we write ${\rm CVaR}_\alpha[F]$ for the CVaR
of a stochastic cost with cumulative distribution function $F$.
\begin{lemma}\label{lemma:cvar-estimation error bound} \cite{wang2022zeroth}  
    Let $F$ and $G$ be two cumulative distribution functions of random variables, within the range $[-U, U]$. Then, 
\begin{equation*} 
        |{\rm CVaR}_\alpha[F] -{\rm CVaR}_\alpha [G] | \le \frac{2U}{\alpha} \sup_{x} \big|F(x)-G(x)\big|.
    \end{equation*}
\end{lemma}

\subsection{Algorithm}
Given the difficulty of estimating the CVaR gradient from a single sample, we sample multiple stochastic costs to estimate it at each iteration.   
At each time slot $t$, the perturbation direction $u_t$ is sampled uniformly from
$\mathbb{S}^{d-1}$. 
The perturbation directions $u_1, \ldots, u_T$ are mutually independent
across time and independent of the sampling noise.
Then, we evaluate the function at $\hat{x}_t = x_t + \delta u_t$.
Denote $m_t$ as the sampling number at time slot $t$.  Let  $m_t  $ satisfy
\begin{align}\label{eq:sampling requirement}
    \sum_{t=1}^{T} \frac{1}{\sqrt{m_t}} \le c T^{1-\frac{a}{2}},
\end{align}
where $0<a<2$, $c>0$ are tuning parameters.  
An example of a sampling policy that satisfies \eqref{eq:sampling requirement} is provided in \cite{wang2022zeroth}, which is given as
\begin{equation}\label{eq:sampling example}
 m_t=   \left\lceil b ( T-t+1)^a\right\rceil,
\end{equation}
where  $b>0$ is the tuning parameter. 
For this illustrative schedule, \(m_t\) is nonincreasing in \(t\). 
In practice, the exact horizon $T$ need not be known and may be
replaced by an estimate of comparable magnitude, which only determines
the scale and decreasing trend of this illustrative schedule.
The parameter $a$ controls the tradeoff between sampling cost and CVaR
estimation accuracy, where a larger $a$ corresponds to a larger sampling
budget and a smaller CVaR estimation error, and vice versa.

We query the stochastic cost $m_t$ times at the perturbed
action, obtaining a batch of samples $\{J_t(\hat{x}_t,\xi_t^i)\}_{i=1}^{m_t}$.
The samples
$\xi_t^1,\ldots,\xi_t^{m_t}$ are i.i.d., and the current batch is
independent of the previously generated sample batches.
At time slot $t$, the learner receives 
\begin{align}
\label{eq:collection one point}
\mathcal{L}_t &= \big \{\big(J_{s_{n|t}}(\hat{x}_{s_{n|t}},\xi_{s_{n|t}}^i), u_{s_{n|t}} \big)\ | \ s_{n|t} + d_{s_{n|t}} = t, \nonumber \\ 
&\hspace{2em}s_{n|t} \le t, i=1,\ldots,m_{s_{n|t}}, n = 1,\dots, \Delta_t \big\}.
\end{align} 
Then, the empirical distribution function is constructed as:
\begin{align}\label{eq:EDF}
    \hat{F}_{s_{n|t}}(y)&=\frac{1}{m_{s_{n|t}}} \sum_{i=1}^{m_{s_{n|t}}} \mathbf{1} \{J_{s_{n|t}}(\hat{x}_{s_{n|t}}, \xi_{s_{n|t}}^i) \leq y \},
\end{align} 
for $n = 1,\dots,\Delta_t$. 
Accordingly, by the CVaR definition \eqref{eq:CVaR def}, 
the CVaR estimate is denoted as ${\rm CVaR}_\alpha[\hat{F}_{s_{n|t}}]$  and the corresponding CVaR  gradient estimate is 
\begin{equation}\label{eq:estimated gradient}
    \hat{g}_t^n=\frac{d}{\delta} {\rm CVaR}_\alpha\big[\hat{F}_{s_{n|t}}\big] u_{s_{n|t}},
\end{equation}
for $n = 1,\dots, \Delta_t$ and $t = 1, \dots, T$. 
For {\color{blue}$0<\delta<r$}, denote $\mathcal{X}^\delta = \{x \in \mathcal{X} \vert \frac{1}{1-\delta/r} x \in \mathcal{X}\}$ as the projection set, and
$\mathcal{P}_{\mathcal{X}^\delta }(x):= {\textrm{argmin}}_{y\in \mathcal{X}^\delta}\|x - y\|^2$ the  projection operation. 
When the learner receives multiple feedback, the gradient descent  evolves as 
\begin{equation}\label{eq:gradient descent}
\begin{array}{rl}
    x_{t}^n & = \mathcal{P}_{\mathcal{X}^\delta}(x_t^{n-1} - \eta \hat{g}_t^n), \ {\rm for}~ n= 1, \dots, \Delta_t, \\
    x_{t+1} & =  x_{t}^{\Delta_t},
\end{array}
\end{equation} 
where $x_{t}^0 = x_t$, and the initial value $x_1\in\mathcal X^\delta$.
When the learner does not receive any feedback at time slot $t$, the risk-averse learning algorithm updates as $x_{t+1} = x_{t}$. 
The projection keeps the actions $\hat{x}_t$ within the admissible set $\mathcal{X}$ as 
$  \big(1-\frac{\delta}{r}\big)\mathcal{X} \oplus \delta   \mathbb{B} = \big(1-\frac{\delta}{r}\big)\mathcal{X} \oplus \frac{\delta}{r} r   \mathbb{B}  \subseteq \big(1-\frac{\delta}{r}\big)\mathcal{X} \oplus \frac{\delta}{r} \mathcal{X} = \mathcal{X}$.
Since $x_1\in\mathcal X^\delta$, every projected update remains in
$\mathcal X^\delta$. Hence, $x_t\in\mathcal X^\delta$ and
$\hat x_t=x_t+\delta u_t\in\mathcal X$ for every $t$.

In summary, the learner utilizes the received feedback $\mathcal{L}_t$ to construct the CVaR gradient estimate and update the decision. In the absence of feedback, the decision remains unchanged. This procedure is outlined in Algorithm~\ref{alg: one-point}. 
Here,  one-point indicates that Algorithm~\ref{alg: one-point} queries a single perturbed action, 
while batch indicates that it draws $m_t$ i.i.d. samples at that action. 
We therefore refer the algorithm as a one-point batch zeroth-order algorithm.
\begin{algorithm}[t] 
\caption{One-point batch risk-averse learning with delayed feedback} \label{alg: one-point}
\begin{algorithmic}[1]
\REQUIRE Initial value $x_1\in\mathcal X^\delta$, iteration horizon $T$, smoothing parameter $\delta$, risk level $\alpha$, step size $\eta$
\FOR{$ {\rm{time}} \;t = 1,\dots, T$} 
\STATE  Sample $u_{t} \sim \mathbb{S}^{d-1}$ and play $\hat{x}_{t}=x_{t}+\delta u_{t} $
\STATE Select sample size $m_t $ 
and query $J_t(\hat{x}_{t},\xi_t^i)$, for $i=1,\ldots,m_t$ 
\STATE Collect feedback $\mathcal{L}_t$, as in \eqref{eq:collection one point}  
\IF{$\mathcal{L}_t = \emptyset$} 
\STATE Update decision:  $x_{t+1} = x_t$
\ELSE 
\STATE Set $x_t^0 = x_t$
\FOR{ $n = 1,\dots, \Delta_t$ }
\STATE Build empirical distribution function $\hat{F}_{s_{n|t}}(y)$, as in \eqref{eq:EDF}
\STATE Estimate CVaR $ {\rm{CVaR}}_{\alpha}[\hat{F}_{s_{n|t}}] $ and construct  CVaR gradient estimate $\hat{g}_{t}^n$, as in \eqref{eq:estimated gradient}
\STATE Set $x_{t}^n = \mathcal{P}_{\mathcal{X}^\delta}(x_t^{n-1} - \eta \hat{g}_t^n)$
\ENDFOR
\STATE Update decision:  $x_{t+1} = x_t^{\Delta_t}$
\ENDIF
\ENDFOR
\end{algorithmic}
\end{algorithm}

\subsection{Regret analysis}
To quantify the error induced by stale feedback,
we introduce a virtual timeline indexed by the feedback utilzation order.
Let $L_t := \sum_{k=1}^t \Delta_k$, with $L_0 := 0,$
denote the total number of feedback batches received by the end of real time slot $t$.
Since some feedback generated near the end of the horizon may arrive after $T$, we have $L_T \le T$.
At real time slot $t$, we have $s_{1|t} < \dots < s_{\Delta_t|t}$. 
This means that the generation times of the feedback batches contained in $\mathcal{L}_t$ are
ordered according to their utilization order. 
Since all feedback arriving at the same time slot is processed within that slot, 
reordering such feedback does not affect the resulting decision sequence, 
but simplifies the subsequent regret analysis.
For $n = 1, \dots, \Delta_t$, define
$\tau = L_{t-1} + n,$ with $\mu(\tau) := s_{n|t}$.
Thus, $\mu(\tau)$ is the generation time of the feedback batch used at
the $\tau$-th virtual time slot. Equivalently,
$$\tilde{J}_\tau(\cdot, \cdot) := J_{\mu(\tau)}(\cdot, \cdot), \quad \tilde{u}_\tau := u_{\mu(\tau)}, \quad \tilde{\xi}_\tau^i := \xi_{\mu(\tau)}^i,$$
and $\tilde{m}_\tau := m_{\mu(\tau)}$. The $\tau$-th utilized batch was queried at
$\widehat{x}_{\mu(\tau)} = x_{\mu(\tau)} + \delta \tilde{u}_\tau,$
whereas it is applied upon receipt to the current virtual iterate
$\tilde{x}_\tau$, defined as the decision immediately before the
$\tau$-th virtual update. Fig.~\ref{fig:virtual mapping} illustrates
this mapping.

In this study, we analyze the regret bound of the algorithm in terms of the cumulative delay and the iteration horizon.
To characterize the cumulative delay,
we introduce an auxiliary variable $\tilde{s}_\tau = \tau - 1 - L_{\scriptsize{\mu(\tau)-1}}$,
which represents the number of virtual updates performed between the generation of $\tau$-th utilized feedback and its utilization.
Hence,
$D_T:=\sum_{\tau=1}^{L_T}\tilde s_\tau$ measures the total delay
that affects the updates within the horizon.
The following lemma is provided to bound $\tilde{s}_\tau$ and its cumulative sum.
\begin{figure}
    \centering
    \scalebox{0.65}{
\begin{tikzpicture}[
  timeline/.style={draw, thick}, 
  dashedtimeline/.style={draw, thick, dashed}, 
  event/.style={draw, fill=blue!20, circle, minimum size=2mm, inner sep=0}, 
  label/.style={font=\small, align=center} 
]

\coordinate (start) at (0,0);
\coordinate (end) at (12,0);
\coordinate (top) at (0,2); 
\coordinate (topend) at (12,2); 

\draw[dashedtimeline] (start) -- (1,0); 
\draw[timeline] (1,0) -- (11,0); 
\draw[dashedtimeline] (11,0) -- (end); 

\draw[dashedtimeline] (top) -- (1,2); 
\draw[timeline] (1,2) -- (11,2); 
\draw[dashedtimeline] (11,2) -- (topend); 

\node[below=1cm, font=\small] at (0.4,3.6) {\textsf{Real}};
\node[event] (event1) at (2,2) {};
\node[label, above=2mm] at (event1) {$\mathcal{L}_{t-2}=\emptyset$};

\node[event] (event2) at (5,2) {};
\node[label, above=2mm] at (event2) {$\mathcal{L}_{t-1}=\emptyset$};

\node[event] (event3) at (8,2) {};
\node[label, above=6mm] at (event3) {$\mathcal{L}_t=\{(J_{t-2},u_{t-2}),$};
\node[label, above=2mm] at (event3) {$(J_{t},u_{t}) \}$};

\node[event] (event4) at (11,2) {};
\node[label, above=6mm] at (event4) {$\mathcal{L}_{t+1} =$};
\node[label, above=2mm] at (event4) {$ \{(J_{t-1},u_{t-1}) \}$};

\node[below=1.2cm, font=\small] at (0.5,1) {\textsf{Virtual}};
\node[above=-1.0cm, font=\small] at (0.5,1) {$\tau = L_{t-1}+1$};
\node[event] (event5) at (2,0) {};
\node[label, below=2mm] at (event5) {$\{(\tilde{J}_{\tau}, \tilde{u}_\tau)\} $}; 
\node[label, below=7mm] at (event5) { $= \{(J_{t-2}, u_{t-2})\}$}; 

\node[event] (event6) at (5,0) {};
\node[label, below=2mm] at (event6) {$\{(\tilde{J}_{\tau+1},\tilde{u}_{\tau+1})\} $};
\node[label, below=7mm] at (event6) {$ =\{(J_{t}, u_{t})\}$};
 
\node[event] (event7) at (8,0) {};
\node[label, below=2mm] at (event7) {$\{(\tilde{J}_{\tau+2},\tilde{u}_{\tau+2})\}$};

\node[event] (event8) at (11,0) {};
\node[label, below=2mm] at (event8) {$\cdots$};

\draw[->] (event3) -- (event5);
\draw[->] (event3) -- (event6);
\draw[->] (event4) -- (event7);

\end{tikzpicture}}
\caption{Real-to-virtual mapping.}
    \label{fig:virtual mapping}
\end{figure}
\begin{lemma}\label{lemma:delay}
By Assumption~\ref{ass:max delay}, the following claims hold: 
(1) $ 0\le \tilde{s}_\tau  \le 2\bar{d}$, for $\tau =\{1,\dots,L_T\}$; 
(2)  Suppose that $T\ge3\bar d+1$.  Denote $D_T:=\sum_{\tau=1}^{L_T}\tilde{s}_\tau$. Then,
$D_T\le\bar D_T $, with $\bar{D}_T = \sum_{t=1}^{L_T-2\bar{d}-1}  d_t + \bar{d}L_T - \frac{\bar{d}^2+\bar{d}}{2}$.
\end{lemma}

The proof of Lemma~\ref{lemma:delay} is provided in the Appendix.  
Some work, e.g., \cite{li2019bandit,wan2023non}, assumes
that all feedback arrives by the end of the horizon $T$, or that the
effect of packets arriving after $T$ is negligible.
In these settings, the
cumulative delay equals
$\sum_{t=1}^{T}d_t$.
Since $\tilde s_\tau\ge0$ and $D_T$ sums only the delays of packets utilized by time $T$, it follows that
$ D_T\le\sum_{t=1}^{T}d_t$.
Moreover, the first claim of
Lemma~\ref{lemma:delay} gives the uniform bound
$D_T\le2\bar dL_T$. 

Before presenting the main results, we define the path length of optimal decisions, also known as the path length of comparators \cite{zhao2021bandit}.
\begin{definition}\label{definition:comparator}
The path length of the comparators is defined as $    P_T = \sum_{t=1}^{T-1}\|x_t^\ast-x_{t+1}^\ast\| $,
where $x_t^\ast = \arg\min_{x \in \mathcal{X}}C_t(x)$ is the optimal decision at time slot $t$. 
\end{definition}

 The regret analysis treats the realized delay sequence as
fixed. At the $\tau$-th virtual update, the arriving packet and its
generation time $\mu(\tau)$ have already been observed and are therefore
deterministic for the learner.  
 For the
regret analysis, let $\mathcal{F}_\tau$ contain the information revealed
before the $\tau$-th virtual update. 
Specifically, define
$\mathcal{F}_1:=\sigma\big(x_1\big)$ and
$\mathcal{F}_{\tau+1}:=
\mathcal{F}_{\tau}\vee \sigma\big(u_{\mu(\tau)},
\xi_{\mu(\tau)}^1,\ldots,\xi_{\mu(\tau)}^{m_{\mu(\tau)}}\big)$,
$\tau=1,\ldots,L_T.$
Thus, $\mathcal F_\tau$ contains the directions and samples of the
first $\tau-1$ utilized packets, but not the direction or samples of
the $\tau$-th packet. 
  
We use $\mathrm{DR}(T)$ to define the generic dynamic regret,
and $\mathrm{DR}_1(T)$ and $\mathrm{DR}_2(T)$ to define the dynamic
regrets of the one- and two-point algorithms, respectively.
The following theorem analyzes the regret of Algorithm~\ref{alg: one-point}. 
\begin{theorem}\label{theorem:one-point}
Let Assumptions \ref{ass:max delay}--\ref{assumption:Lipschitz} hold.
Let the sampling numbers $m_t$ over the horizon $T$ satisfy
\eqref{eq:sampling requirement} and the smoothing radius satisfies
$0<\delta<r$.
For any $\gamma\in(0,1)$, we have the following claims.
\begin{enumerate}
\item When $0<a<1$, selecting $\delta=T^{-a/4}\sqrt{1+\bar dP_T}$ and $\eta=(T+D_T)^{-3a/4}\sqrt{1+\bar dP_T}$, Algorithm~\ref{alg: one-point} achieves
 $\mathrm{DR}_1(T)=\tilde{\mathcal O}\!\big( 
\bigl(T^{1-a/4}+(T+D_T)^{3a/4}\bigr)
\sqrt{1+\bar dP_T} +(T+D_T)^{-3a/4}T^{a/4}D_T+\bar d\big)$
with probability at least $1-\gamma$.
\item When $a\ge1$, selecting $\delta=T^{-1/4}\sqrt{1+\bar dP_T}$ and $\eta=(T+D_T)^{-3/4}\sqrt{1+\bar dP_T}$, Algorithm~\ref{alg: one-point} achieves $
\mathrm{DR}_1(T)=\tilde{\mathcal O}\!\left((T+D_T)^{3/4}\sqrt{1+\bar dP_T}+\bar d\right)$
with probability at least $1-\gamma$.
\end{enumerate}
\end{theorem}
\textit{Proof}.
The regret \eqref{eq:dynamic regret definition} satisfies
\begin{align}\label{eq:convex11}
&\mathrm{DR}_1(T)=\sum_{t=1}^T\bigl(C_t(\hat x_t)-C_t(x_t^\ast)\bigr)\nonumber\\
&\leq \sum_{t=1}^T\bigl(C_t(x_t)-C_t(x_t^\ast)\bigr)
+\delta L_0T\nonumber\\
&\leq \sum_{t=1}^T\bigl(C_t^\delta(x_t)-C_t(x_t^\ast)\bigr)
+2\delta L_0T,
\end{align}
where the first inequality follows from
$\|\hat x_t-x_t\|=\delta$ and the Lipschitz continuity of $C_t$, 
and the second follows from
$|C_t^\delta(x_t)-C_t(x_t)|\le\delta L_0$. Notice that the smoothed
function is evaluated at $x_t\in\mathcal X^\delta$.
For each $t$, define the scaled comparator $    x_t^{\delta,\ast}:=\left(1-\delta/r\right)x_t^\ast $, 
which is a feasible comparator for the projected gradient update. 
Since $x_t^\ast\in\mathcal{X}$, it follows that
$x_t^{\delta,\ast}\in\mathcal{X}^\delta$. 
Moreover, Lemma~\ref{lemma:smoothed cvar Lipschitz} and the Lipschitz
continuity of $C_t$ give
\begin{align}\label{eq:projection mismatch}
&C_t^\delta(x_t^{\delta,\ast})-C_t(x_t^\ast) \le \bigl|C_t^\delta(x_t^{\delta,\ast})
-C_t(x_t^{\delta,\ast})\bigr|
+L_0\|x_t^{\delta,\ast}-x_t^\ast\|\nonumber\\
&\le \left(1+\frac{D_x}{r}\right)\delta L_0,
\end{align}
where the last inequality follows from
$\|x_t^{\delta,\ast}-x_t^\ast\|
=(\delta/r)\|x_t^\ast\|\le \delta D_x/r$.
Substituting \eqref{eq:projection mismatch} into \eqref{eq:convex11}, we have 
\begin{equation}\label{eq:convex1}
\mathrm{DR}_1(T)  \leq \sum_{t=1}^T C_t^\delta(x_t)- 
 \sum_{t=1}^T C_t^\delta (x_t^{\delta,\ast})+\Big(3 +\frac{D_x}{r}\Big)\delta L_0 T.
\end{equation}

At virtual time $\tau$, the utilized packet was generated at time
$\mu(\tau)$. Hence, the first and second terms of \eqref{eq:convex1} can be further written as  
\begin{align}\label{eq:virtual mapping}
  &  \sum_{t=1}^T  C_t^\delta(x_t)-  \sum_{t=1}^T C_t^\delta (x_t^{\delta,\ast}) \nonumber \\ 
 &\leq \sum_{\tau=1}^{L_T}  C_{\mu(\tau)}^\delta(x_{\mu(\tau)})-  \sum_{\tau=1}^{L_T} C_{\mu(\tau)}^\delta (x_{\mu(\tau)}^{\delta,\ast}) + 2U\bar{d}\nonumber  \\ 
  & \leq \sum_{\tau=1}^{L_T} \big\langle \nabla C_{\mu(\tau)}^\delta(x_{\mu(\tau)}),      x_{\mu(\tau)} - x_{\mu(\tau)}^{\delta,\ast}\big\rangle +  2U\bar{d} \nonumber  \\
  & = \sum_{\tau=1}^{L_T} \big\langle \nabla C_{\mu(\tau)}^\delta(x_{\mu(\tau)}),
     x_{\mu(\tau)}-\tilde{x}_\tau+\tilde{x}_\tau-x_{\mu(\tau)}^{\delta,\ast}\big\rangle
      +  2U\bar{d}, \nonumber \\
    \hspace{-2em}   &= R_1 + R_2 + 2U\bar{d},
\end{align}
with
$R_1 = \sum_{\tau=1}^{L_T}\big\langle \nabla C_{\mu(\tau)}^\delta(x_{\mu(\tau)}),
x_{\mu(\tau)}-\tilde{x}_\tau\big\rangle$
and
$R_2 = \sum_{\tau=1}^{L_T}\big\langle \nabla C_{\mu(\tau)}^\delta(x_{\mu(\tau)}),
\tilde{x}_\tau-x_{\mu(\tau)}^{\delta,\ast}\big\rangle$.
The first inequality follows from $T-L_T\le\bar{d}$ and $|C_t^\delta(\cdot)|\le U$. The second inequality follows from the convexity of $C_t^\delta(\cdot)$ for all $t$. In the following, we bound $R_1$ and $R_2$, respectively.

Denote $\tilde{g}_\tau$ as the $\tau$-th CVaR gradient estimate being used for gradient descent \eqref{eq:gradient descent}, for $\tau = 1,\dots, L_T$. 
Since the real iterate $x_t$ incorporates all feedback received by the end of time slot $t - 1$,
it satisfies
$x_t = \tilde{x}_{L_{t-1}+1}.$
From the definition of
$\tilde{s}_\tau $, it follows that
$x_{\mu(\tau)} = \tilde{x}_{\tau - \tilde{s}_\tau}.$
Therefore, we have
\begin{align}\label{eq:delay state mismatch}
& \|x_{\mu(\tau)} - \tilde{x}_\tau\| \nonumber \\
&= \| \tilde{x}_{\tau - \tilde{s}_\tau} - \tilde{x}_\tau \| \le \sum_{j=0}^{\tilde{s}_\tau-1}\| \tilde{x}_{\tau-\tilde{s}_\tau+j} -\tilde{x}_{\tau-\tilde{s}_\tau+j+1} \| \nonumber \\
&\le \sum_{j=0}^{\tilde{s}_\tau-1} \eta \|\tilde{g}_{\tau-\tilde{s}_\tau+j}\| \le \frac{d\eta}{\delta}U\tilde{s}_\tau,  
\end{align}
where the second inequality follows from  $\|\tilde{x}_i - \tilde{x}_{i+1}\| = \|\tilde{x}_i - \mathcal{P}_{\mathcal{X}^\delta}(\tilde{x}_i -\eta \tilde{g}_i)\| \le \eta \| \tilde{g}_i \|  $, and  the last inequality follows from $\|\tilde{g}_i\| \le \frac{dU}{\delta}$ by its definition \eqref{eq:estimated gradient}. 
Thus, for $R_1$ in \eqref{eq:virtual mapping}, we have 
\begin{align}\label{eq:multi-update order}
&R_1\le L_0\sum_{\tau=1}^{L_T}\|x_{\mu(\tau)}-\tilde x_\tau\|\nonumber\\
&\le\frac{\eta L_0dU}{\delta}\sum_{\tau=1}^{L_T}\tilde s_\tau
=\frac{\eta L_0dUD_T}{\delta},
\end{align}
where the first inequality follows from Lemma~\ref{lemma:smoothed cvar Lipschitz}, 
the second inequality follows from \eqref{eq:delay state mismatch}, and the equality follows from the definition of $D_T$.

The next step is to bound $R_2$ in \eqref{eq:virtual mapping}.
By the gradient descent rule \eqref{eq:gradient descent}, we have 
\begin{align*}
  &   \|\tilde{x}_{\tau+1} - x_{\mu(\tau)}^{\delta,\ast} \|^2 \\
 &= \|\mathcal{P}_{\mathcal{X}^\delta }(\tilde{x}_{\tau}-\eta \tilde{g}_\tau) - x_{\mu(\tau)}^{\delta,\ast} \|^2  \\
& \leq  \| \tilde{x}_{\tau}-\eta \tilde{g}_\tau  - x_{\mu(\tau)}^{\delta,\ast}\|^2   \\
& = \| \tilde{x}_\tau-x_{\mu(\tau)}^{\delta,\ast}\|^2+\eta^2\|\tilde{g}_\tau\|^2-2\eta \langle \tilde{g}_\tau,  \tilde{x}_\tau-x_{\mu(\tau)}^{\delta,\ast} \rangle   , 
\end{align*}
where the inequality follows from the fact that $x_{\mu(\tau)}^{\delta,\ast}\in\mathcal{X}^{\delta}$.
Then, we obtain
\begin{align}
\label{eq:gt-x}
&\langle \tilde{g}_\tau,
\tilde{x}_\tau-x_{\mu(\tau)}^{\delta,\ast}\rangle \nonumber \\
&\leq \frac{\|\tilde{x}_\tau-x_{\mu(\tau)}^{\delta,\ast}\|^2-\|\tilde{x}_{\tau+1}-x_{\mu(\tau)}^{\delta,\ast}\|^2}{2\eta}
+\frac{\eta}{2}\|\tilde{g}_\tau\|^2.
\end{align}
Note that constructing the empirical distribution function of the cost function using finite samples induces CVaR estimate error $ \hat{\epsilon}_t:= {\mathrm{CVaR}}_\alpha[\hat{F}_t] - {\mathrm{CVaR}}_\alpha[F_t]$, where $\hat{F}_t$ is the empirical distribution function that corresponds to $F_t$, for $t = 1,\dots, T$. 
Define the ideal estimator constructed from the true distribution as
\begin{equation}\label{eq:ideal one-point estimator}
\tilde g_\tau^\ast
:=\frac d\delta C_{\mu(\tau)}
\bigl(x_{\mu(\tau)}+\delta\tilde u_\tau\bigr)\tilde u_\tau.
\end{equation} 
The virtual iterate $\tilde{x}_\tau$ is a function
of the first $\tau-1$ updates. 
Therefore, the current direction
$\tilde u_\tau=u_{\mu(\tau)}$ is uniform on $\mathbb S^{d-1}$ and
independent of $\mathcal F_\tau$. 
Moreover, $x_{\mu(\tau)}$ depends only on
packets utilized before real time $\mu(\tau)$, all of which precede the
$\tau$-th packet in the virtual order. 
Hence, $x_{\mu(\tau)}$ and
$\tilde{x}_\tau$ are $\mathcal F_\tau$-measurable. Therefore, \eqref{eq:ideal one-point estimator} give
\begin{equation}\label{eq:conditional ideal gradient}
\mathbb E[\tilde g_\tau^\ast | \mathcal F_\tau]
=\nabla C_{\mu(\tau)}^\delta(x_{\mu(\tau)}).
\end{equation}
Moreover, $\tilde g_\tau^\ast=\tilde g_\tau-(d/\delta)\tilde\epsilon_\tau\tilde u_\tau$, 
where $\tilde\epsilon_\tau:=\hat\epsilon_{\mu(\tau)}$.
Define $\zeta_\tau = \tilde{g}_\tau^\ast -\nabla C_{\mu(\tau)}^\delta(x_{\mu(\tau)}) $.
Since $\tilde{u}_\tau \in \mathcal{F}_{\tau+1}$, the term $\zeta_\tau$ is $\mathcal F_{\tau+1}$-measurable and
$\mathbb E[\zeta_\tau | \mathcal F_\tau]=0$. Likewise, the empirical
estimate $\tilde g_\tau$ and its estimation error $\tilde\epsilon_\tau$
are $\mathcal F_{\tau+1}$-measurable.
Since $\|\tilde x_\tau-x_{\mu(\tau)}^{\delta,\ast}\|\le D_x$, $R_2$ can be written as
\begin{align}\label{eq:R2 decomposition}
&R_2 = \sum_{\tau=1}^{L_T}
\left\langle  \tilde g_\tau-
\frac d\delta\tilde\epsilon_\tau\tilde u_\tau - \zeta_\tau, \tilde{x}_\tau-x_{\mu(\tau)}^{\delta,\ast}
\right\rangle \nonumber \\
& \le \sum_{\tau=1}^{L_T}
\left\langle  \tilde g_\tau,
\tilde{x}_\tau-x_{\mu(\tau)}^{\delta,\ast} \right\rangle
+ \frac{dD_x}{\delta}\sum_{\tau=1}^{L_T}|\tilde{\epsilon}_\tau|
\nonumber\\
&\quad- \sum_{\tau=1}^{L_T}\left\langle \zeta_\tau,
\tilde{x}_\tau-x_{\mu(\tau)}^{\delta,\ast} \right\rangle.
\end{align}
Denote $R_{21} = \frac{1}{{2\eta}}\sum_{\tau = 1}^{L_T} \big(  \|\tilde{x}_\tau-x_{\mu(\tau)}^{\delta,\ast}\|^2 -\|\tilde{x}_{\tau+1}-x_{\mu(\tau)}^{\delta,\ast}\|^2 \big) $, $R_{22} =  \frac{dD_x}{\delta}\sum_{\tau=1}^{L_T}|\tilde{\epsilon}_\tau|$, 
and $R_{23} = -\sum_{\tau=1}^{L_T}\left\langle \zeta_\tau, \tilde{x}_\tau-x_{\mu(\tau)}^{\delta,\ast} \right\rangle$. 
Substituting \eqref{eq:gt-x} into the inequality, we obtain
\begin{align}\label{eq:estimated gradient sum}
&\sum_{\tau=1}^{L_T}
\left\langle  \tilde g_\tau, \tilde{x}_\tau-x_{\mu(\tau)}^{\delta,\ast} \right\rangle
\le R_{21}  + \frac{\eta}{2} \sum_{\tau=1}^{L_T}
\|\tilde{g}_\tau \|^2 \nonumber \\ 
&\le  R_{21} + \frac{\eta d^2U^2}{2\delta^2}L_T.
\end{align}
Combining \eqref{eq:R2 decomposition} and
\eqref{eq:estimated gradient sum} gives
\begin{equation}\label{eq:virtual order}
R_2\le R_{21}+R_{22}+R_{23}
+\frac{\eta d^2U^2}{2\delta^2}L_T.
\end{equation}
 
Furthermore, we have
\begin{align}\label{eq:R21}
&R_{21} = \frac{1}{2\eta}\sum_{\tau =1}^{L_T}
\left(\|\tilde{x}_\tau\|^2-\|\tilde{x}_{\tau+1}\|^2
+2(x_{\mu(\tau)}^{\delta,\ast})^\top(\tilde{x}_{\tau+1}-\tilde{x}_\tau)\right) \nonumber \\
&\le \frac{1}{2\eta}\Big(D_x^2
+2\sum_{\tau=2}^{L_T}\tilde{x}_\tau^\top
(x_{\mu(\tau-1)}^{\delta,\ast}-x_{\mu(\tau)}^{\delta,\ast}) \nonumber \\
&\hspace{1em}
+2(x_{\mu(L_T)}^{\delta,\ast})^\top\tilde{x}_{L_T+1}
-2(x_{\mu(1)}^{\delta,\ast})^\top\tilde{x}_1\Big) \nonumber \\
&\le \frac{1}{2\eta}\left(
5D_x^2+2D_x\sum_{\tau=2}^{L_T}
\|x_{\mu(\tau-1)}^{\delta,\ast}-x_{\mu(\tau)}^{\delta,\ast}\|
\right) \nonumber \\
&\le \frac{1}{2\eta}\left(
5D_x^2+2D_x\sum_{\tau=2}^{L_T}
\|x_{\mu(\tau-1)}^{\ast}-x_{\mu(\tau)}^{\ast}\|
\right) \nonumber \\
&\le \frac{5D_x^2+4D_x\bar dP_T}{2\eta},
\end{align}
where the first and the second inequalities follow from the fact that $D_x$ is the diameter of $\mathcal{X}$. The third inequality follows from the definition of the scaled comparators, which gives
$\|x_{\mu(\tau-1)}^{\delta,\ast}-x_{\mu(\tau)}^{\delta,\ast}\|
=(1-\delta/r)\|x_{\mu(\tau-1)}^{\ast}-x_{\mu(\tau)}^{\ast}\|
\le\|x_{\mu(\tau-1)}^{\ast}-x_{\mu(\tau)}^{\ast}\|$. 
The last inequality follows from  Lemma~\ref{lemma:comparator}.  

We next bound $R_{22}$. 
At virtual time $\tau$, $x_{\mu(\tau)}$ is
$\mathcal F_\tau$-measurable, while the current direction
$\tilde u_\tau$ and the current sample batch are not contained in
$\mathcal F_\tau$. Conditional on
$\mathcal F_\tau\vee\sigma(\tilde u_\tau)$, the queried action and
direction are fixed and the $\tilde m_\tau$ samples in the current batch
are i.i.d. Therefore, the Dvoretzky--Kiefer--Wolfowitz (DKW) inequality
\cite{dvoretzky1956asymptotic} gives, for every utilized packet, 
\begin{equation}\label{eq:DKW}
\begin{split}
&\mathbb P\!\left\{
\sup_y|\hat F_{\mu(\tau)}(y)-F_{\mu(\tau)}(y)|
\ge\sqrt{\frac{\ln(4T/\gamma)}{2\tilde m_\tau}}
\right.\\[-0.2em]
&\hspace{8em}\left.
\middle|\ \mathcal F_\tau\vee\sigma(\tilde u_\tau)\right\}
\le\frac{\gamma}{2T}.
\end{split}
\end{equation} 
 Denote the event in \eqref{eq:DKW} by $A_\tau$. 
Taking
expectations gives
$
\mathbb P(A_\tau)
=\mathbb E\!\left[
\mathbb P\!\left(A_\tau\mid
\mathcal F_\tau\vee\sigma(\tilde u_\tau)\right)\right]
\le\frac{\gamma}{2T}.
$
Since $L_T\le T$, a union bound over the utilized packets gives
$
\mathbb P\!\left\{\exists\,\tau\le L_T:A_\tau\right\}
\le T\frac{\gamma}{2T}
=\frac{\gamma}{2}.
$
Together with Lemma~\ref{lemma:cvar-estimation error bound}, 
\begin{equation}\label{eq:DKW cvar mismatch}
|\tilde\epsilon_\tau|\le\frac{2U}{\alpha}
\sqrt{\frac{\ln(4T/\gamma)}{2\tilde m_\tau}},
\qquad \tau=1,\ldots,L_T.
\end{equation}
This bound holds simultaneously for all
$\tau=1,\ldots,L_T$ with probability at least $1-\gamma/2$.
Consequently, \eqref{eq:sampling requirement} yields
\begin{align}\label{eq:estimation error order}
&R_{22}
\le\frac{2dUD_x}{\alpha\delta}
\sqrt{\frac{\ln(4T/\gamma)}{2}}
\sum_{\tau=1}^{L_T}\frac{1}{\sqrt{\tilde m_\tau}}\nonumber\\
&\le\frac{2cdUD_x}{\alpha\delta}
\sqrt{\frac{\ln(4T/\gamma)}{2}}T^{1-a/2}.
\end{align}

We next bound $R_{23}$. 
By \eqref{eq:conditional ideal gradient} and the
$\mathcal F_\tau$-measurability of
$\tilde x_\tau-x_{\mu(\tau)}^{\delta,\ast}$,
$
\mathbb E[\langle\zeta_\tau,
\tilde x_\tau-x_{\mu(\tau)}^{\delta,\ast}\rangle | \mathcal F_\tau]=0.
$ 
Moreover,
$
\left|\langle\zeta_\tau,
\tilde x_\tau-x_{\mu(\tau)}^{\delta,\ast}\rangle\right|
\le \|\zeta_\tau\|D_x
\le 2dUD_x/\delta,
$
where the last inequality follows from
$\|\tilde g_\tau^\ast\|\le dU/\delta$ and
$\|\nabla C_{\mu(\tau)}^\delta(x_{\mu(\tau)})\|
\le dU/\delta$. 
Since $\zeta_\tau$ is $\mathcal F_{\tau+1}$-measurable,
the summands of $R_{23}$ are conditionally centered bounded increments.
Therefore, the Azuma--Hoeffding inequality gives, for every $s>0$,
\begin{equation*}
\mathbb P(R_{23}\ge s)
\le \exp\!\left(-\frac{s^2}{2L_T
\left(2dUD_x/\delta\right)^2}\right).
\end{equation*}
Taking $s=(2dUD_x/\delta)\sqrt{2L_T\ln(2/\gamma)}$ gives, with
probability at least $1-\gamma/2$,
\begin{equation}\label{eq:R23}
R_{23}\le\frac{2dUD_x}{\delta}
\sqrt{2L_T\ln\frac{2}{\gamma}}
\le\frac{2dUD_x}{\delta}
\sqrt{2T\ln\frac{2}{\gamma}}.
\end{equation}
Combining \eqref{eq:convex1}, \eqref{eq:virtual mapping},
\eqref{eq:multi-update order}, \eqref{eq:virtual order},
\eqref{eq:R21},
\eqref{eq:estimation error order}, and
\eqref{eq:R23}, and applying a union bound to the last two
high-probability events, we obtain
\begin{align}\label{eq:one-point nonasymptotic bound}
&\mathrm{DR}_1(T)
\le (3+D_x/r)\delta L_0T+2U\bar d +\frac{\eta L_0dU}{\delta}D_T
\nonumber\\
&\hspace{0.5em}+\frac{\eta d^2U^2}{2\delta^2}T+\frac{5D_x^2+4D_x\bar dP_T}{2\eta}\nonumber\\
&\hspace{0.5em}+\frac{2cdUD_x}{\alpha\delta}
\sqrt{\frac{\ln(4T/\gamma)}{2}}T^{1-a/2} +\frac{2dUD_x}{\delta}\sqrt{2T\ln\frac{2}{\gamma}},
\end{align}
with probability at least $1-\gamma$.
Substituting the choices of $\delta$ and
$\eta$ from the theorem into \eqref{eq:one-point nonasymptotic bound}, we obtain the result.
\hfill $\blacksquare$
\begin{remark}\label{remark:clean bound}
The two cases in Theorem~\ref{theorem:one-point} admit the
following unified conservative form:
\[
\mathrm{DR}_1(T)=\widetilde{\mathcal O}\!\left(
(T+D_T)^{1-\frac14\min\{1,a\}}
\sqrt{1+\bar dP_T}+\bar d\right).
\]
For $0<a<1$, the horizon exponent $1-a/4$ decreases as the sampling
parameter $a$ increases, whereas for $a\ge1$ it saturates at $3/4$.
The interaction between delay and temporal variation is inherent to
learning with delayed feedback in a changing environment.
Specifically, delay may reorder the feedback packets, and Lemma~\ref{lemma:comparator} shows
that each increment of the original comparator path can be counted at
most $2\bar d$ times in the reordered sequence. Consequently, the
cumulative comparator drift is bounded by $2\bar dP_T$. 
However, the product
$\bar dP_T$ follows from our worst-case reordering argument, and the
present analysis does not establish that this dependence is minimax
optimal.
When $P_T=0$, this coupling term disappears, although the
delay-dependent terms involving $D_T$ and $\bar d$ remain. 

In implementation, the parameters $P_T$, $\bar d$, and $D_T$ may be
unknown and can be replaced by deterministic upper bounds. Using larger
upper bounds leads to more conservative parameter choices and
performance guarantees. In particular, an overestimate of $\bar d$
preserves the delay assumption. If $\bar d$ is underestimated and an
actual delay exceeds it, the current theoretical guarantees no longer
apply. Finally, the dimension dependence in
\eqref{eq:one-point nonasymptotic bound} originates from the standard
spherical zeroth-order gradient estimator. This dependence is standard
for this estimator under Euclidean geometry, but we do not claim that
it is minimax optimal in the dimension.
\end{remark}
\begin{remark}\label{remark:compare zhao}
In the delay-free setting, the one-point bandit method in
\cite{zhao2021bandit} achieves
$\widetilde{\mathcal O}(T^{3/4}\sqrt{1+P_T})$. When $a\ge1$ and delay is
absent, the bound in Remark~\ref{remark:clean bound} has the same
horizon and path-length order.
When $P_T$ is unknown, \cite{zhao2021bandit} uses a meta-algorithm that
combines candidate learners to select the best-performing one
adaptively. Extending this method to the present delayed
risk-averse setting would require coordinating delayed updates and
multiple empirical CVaR estimators, and is left for future work. 
 
\end{remark}
\begin{remark}\label{remark:technical distinction}
Compared with existing delayed online convex optimization
methods for deterministic loss functions \cite{wan2023non}, 
the feedback in our setting contains one or two batches of stochastic
function evaluations, which are used to form an empirical CVaR gradient
estimate. We therefore define the filtrations according to the virtual
utilization order to justify the conditional expectation and
concentration arguments.
Moreover, unlike in delay-free zeroth-order CVaR learning, the
estimation errors in our setting are carried by delayed feedback packets,
and contribute to the stale iterate mismatch through delayed
utilization in the regret analysis.
Our regret analysis explicitly
quantifies these coupling effects and controls them through the sampling
requirement~\eqref{eq:sampling requirement}, together with the
cumulative- and maximum-delay bounds.
\end{remark}

\section{Two-point risk-averse learning}\label{sec:two-point}
This section develops a two-point zeroth-order risk-averse learning
algorithm. Compared with Algorithm~\ref{alg: one-point}, the
learner uses two independent sample batches, one at each symmetric
perturbation, to form the gradient estimate. 

\subsection{Algorithm}
At each time slot $t$, sample $u_t$ uniformly from $\mathbb S^{d-1}$,
independently across time and independently of the sampling noise, and
query the two perturbed actions
$
\hat x_t^1=x_t+\delta u_t$, $\hat x_t^2=x_t-\delta u_t$. 
For the two perturbed actions, use two mutually independent sample
batches $\{\xi_t^{i,1}\}_{i=1}^{m_t}$ and
$\{\xi_t^{i,2}\}_{i=1}^{m_t}$. 
The samples within each batch are i.i.d.,
and are independent of the previous sample
batches and of $u_t$.
We assume that the paired sample batches experience the same delay.
Over the horizon, the sampling numbers satisfy
\eqref{eq:sampling requirement}.
At time slot $t$, the learner receives
\begin{align}\label{eq:two point feedback collection}
&\bar{\mathcal L}_t
=\big\{\big(J_{s_{n|t}}(\hat x_{s_{n|t}}^1,
\xi_{s_{n|t}}^{i,1}), J_{s_{n|t}}(\hat x_{s_{n|t}}^2,
\xi_{s_{n|t}}^{i,2}),u_{s_{n|t}}\big)\ \big|\nonumber\\
&\hspace{0.5em} s_{n|t}+d_{s_{n|t}}=t,
  n=1,\ldots,\Delta_t, i=1,\ldots,m_{s_{n|t}}\big\}.
\end{align}
Define the two empirical distribution functions:
\begin{align}\label{eq:EDF 2}
\hat F_{s_{n|t}}^1(y)
&=\frac{1}{m_{s_{n|t}}}
\sum_{i=1}^{m_{s_{n|t}}}
\mathbf 1\!\left\{
J_{s_{n|t}}(\hat x_{s_{n|t}}^1,\xi_{s_{n|t}}^{i,1})\le y
\right\},\nonumber\\
\hat F_{s_{n|t}}^2(y)
&=\frac{1}{m_{s_{n|t}}}
\sum_{i=1}^{m_{s_{n|t}}}
\mathbf 1\!\left\{
J_{s_{n|t}}(\hat x_{s_{n|t}}^2,\xi_{s_{n|t}}^{i,2})\le y
\right\}.
\end{align}
For $n=1,\ldots,\Delta_t$, the two-point CVaR gradient
estimate is
\begin{equation}\label{eq:estimated gradient 2}
\bar g_t^n
=\frac{d}{2\delta}
\left(
{\rm CVaR}_\alpha[\hat F_{s_{n|t}}^1]
-{\rm CVaR}_\alpha[\hat F_{s_{n|t}}^2]
\right)u_{s_{n|t}}.
\end{equation}
The received batches are processed in increasing order of their
generation times, same as in Algorithm~\ref{alg: one-point}. 
The two-point zeroth-order algorithm is outlined in Algorithm~\ref{alg: two-point}. At each iteration, Algorithm~\ref{alg: two-point} queries two perturbed actions, 
and draws $m_t$ i.i.d. samples at each action, resulting in two batches of $m_t$ stochastic function evaluations. 
\begin{algorithm}[t]
\caption{Two-point batch risk-averse learning with delayed feedback}
\label{alg: two-point}
\begin{algorithmic}[1]
\REQUIRE Initial value $x_1\in\mathcal X^\delta$, iteration horizon
$T$, smoothing parameter $ \delta $, risk level $\alpha$, step size
$\eta$ 
\FOR{$t=1,\ldots,T$}
\STATE Sample $u_t\sim\mathbb S^{d-1}$ and play
$\hat x_t^1=x_t+\delta u_t$ and $\hat x_t^2=x_t-\delta u_t$
\STATE Query $J_t(\hat x_t^1,\xi_t^{i,1})$ and
$J_t(\hat x_t^2,\xi_t^{i,2})$,
$i=1,\ldots,m_t$
\STATE Collect feedback $\bar{\mathcal L}_t$ as in
\eqref{eq:two point feedback collection}
\IF{$\bar{\mathcal L}_t=\emptyset$}
\STATE Set $x_{t+1}=x_t$
\ELSE
\STATE Set $x_t^0=x_t$
\FOR{$n=1,\ldots,\Delta_t$}
\STATE Build $\hat F_{s_{n|t}}^1$ and $\hat F_{s_{n|t}}^2$ as in
\eqref{eq:EDF 2}
\STATE Construct $\bar g_t^n$ as in \eqref{eq:estimated gradient 2}
\STATE Set
$x_t^n=\mathcal P_{\mathcal X^\delta}(x_t^{n-1}-\eta\bar g_t^n)$
\ENDFOR
\STATE Set $x_{t+1}=x_t^{\Delta_t}$
\ENDIF
\ENDFOR
\end{algorithmic}
\end{algorithm}

\subsection{Regret analysis}
Since the learner queries two perturbed actions per iteration, we
define the two-point dynamic regret as
\begin{equation}\label{eq:DR 2}
\mathrm{DR}_2(T)
=\sum_{t=1}^T\frac{C_t(\hat x_t^1)+C_t(\hat x_t^2)}{2}
-\sum_{t=1}^T C_t(x_t^\ast).
\end{equation}
We use the virtual timeline $L_t$, the mapping $\mu(\tau)$, the virtual
iterate $\tilde x_\tau$, the virtual delay
$\tilde s_\tau$, the cumulative virtual delay
$D_T=\sum_{\tau=1}^{L_T}\tilde s_\tau$, the virtual sampling number
$\tilde m_\tau$ introduced in Section~\ref{sec:one-point}. Because each
two-point packet contains two sample batches, define the filtration
\begin{align*}
&\bar{\mathcal F}_{\tau+1},
:=\bar{\mathcal F}_\tau\vee\sigma\!\left(
u_{\mu(\tau)},\{\xi_{\mu(\tau)}^{i,1}\}_{i=1}^{m_{\mu(\tau)}},
 \xi_{\mu(\tau)}^{i,2}\}_{i=1}^{m_{\mu(\tau)}}\right),\nonumber\\
&\bar{\mathcal F}_1
:=\sigma(x_1), 
\quad \tau=1,\ldots,L_T.
\end{align*}
Thus, $\bar{\mathcal F}_\tau$ contains the directions and both sample
batches of the first $\tau-1$ utilized packets, but not those of the
$\tau$-th packet. In particular, $x_{\mu(\tau)}$, $\tilde x_\tau$, and
$x_{\mu(\tau)}^{\delta,\ast}$ are
$\bar{\mathcal F}_\tau$-measurable, whereas
$\tilde u_\tau=u_{\mu(\tau)}$ is conditionally uniform on
$\mathbb S^{d-1}$ and independent of $\bar{\mathcal F}_\tau$.
For the two-point algorithm, let $\tilde{\bar g}_\tau$ denote the
gradient estimate used at the $\tau$-th virtual update.

The following theorem analyzes Algorithm~\ref{alg: two-point}.
\begin{theorem} 
\label{theorem:two-point}
Let Assumptions~\ref{ass:max delay}--\ref{assumption:Lipschitz} hold,
and let the two perturbations use mutually independent i.i.d. sample
batches.  
Let the sampling numbers $m_t$ over the horizon $T$ satisfy
\eqref{eq:sampling requirement} and the smoothing radius satisfies
$0<\delta<r$.
Select
$
\delta=T^{-a/4}\sqrt{1+\bar dP_T}$,
$\eta=(T+D_T)^{-1/2}\sqrt{1+\bar dP_T}.$
Assume that the stated smoothing radius satisfies $0<\delta<r$.
 Then, for any $\gamma\in(0,1)$, and $P_T\ge1$,  Algorithm~\ref{alg: two-point}
achieves, with probability at least $1-\gamma$,  
\begin{equation}\label{eq:two-point simplified path bound}
\mathrm{DR}_2(T)=\widetilde{\mathcal O}\!\left(
\left(T^{1-a/4}+(T+D_T)^{1/2}\right)
\sqrt{1+\bar dP_T}+\bar d\right).
\end{equation}
\end{theorem} 
\textit{Proof}. 
By the $L_0$-Lipschitz continuity of $C_t$, for every realization of $u_t$,
\begin{align}
&\frac{1}{2}\big(  C_t(x_t+\delta u_t)+C_t(x_t-\delta u_t)\big)
\nonumber\\ 
&\le C_t(x_t)+\delta L_0\le C_t^\delta(x_t)+2\delta L_0.
\label{eq:two-point Lipschitz reduction}
\end{align}
By \eqref{eq:DR 2} and
Lemma~\ref{lemma:smoothed cvar Lipschitz},
\begin{align}\label{eq:DR two-point}
&\mathrm{DR}_2(T)
\le \sum_{t=1}^T
\big(C_t^\delta(x_t)-C_t(x_t^\ast)\big)
+2\delta L_0T\nonumber\\
&\le \sum_{t=1}^T
\big(C_t^\delta(x_t)-C_t^\delta(x_t^{\delta,\ast})\big)
+\left(3+\frac{D_x}{r}\right)\delta L_0T.
\end{align}
The second inequality in \eqref{eq:DR two-point} follows from the
scaled comparator $x_t^{\delta,\ast}=(1-\delta/r)x_t^\ast$ and
\eqref{eq:projection mismatch}.
  
Following the real-to-virtual decomposition used in the proof of Theorem~\ref{theorem:one-point}, 
we obtain 
\begin{equation}\label{eq:two-point virtual decomposition}
\mathrm{DR}_2(T)
\le R_1+R_2+2U\bar d
+\left(3+\frac{D_x}{r}\right)\delta L_0T,
\end{equation} 
where $R_1$ and $R_2$ are defined in
\eqref{eq:virtual mapping}.

We first bound the empirical CVaR errors. For
$\ell\in\{1,2\}$, let $F_t^\ell$ be the true distribution function of
$J_t(\hat x_t^\ell,\xi_t)$ and define
$
\epsilon_t^\ell
:={\rm CVaR}_\alpha[\hat F_t^\ell]
-{\rm CVaR}_\alpha[F_t^\ell]$,
$\bar\epsilon_t:=\epsilon_t^1-\epsilon_t^2,$
$\tilde{\bar\epsilon}_\tau:=\bar\epsilon_{\mu(\tau)}$.
Equivalently,
\begin{align}\label{eq:empirical error decomposition 2}
\bar\epsilon_t
&={\rm CVaR}_\alpha[\hat F_t^1]
-{\rm CVaR}_\alpha[\hat F_t^2]\nonumber\\
&\quad-\left({\rm CVaR}_\alpha[F_t^1]
-{\rm CVaR}_\alpha[F_t^2]\right).
\end{align} 
At virtual time $\tau$, conditional on
$\bar{\mathcal F}_\tau\vee\sigma(\tilde u_\tau)$, the queried actions
are fixed, the two current sample batches are mutually independent, and
the samples within each batch are i.i.d. Hence, for every utilized
packet and $\ell\in\{1,2\}$, the DKW inequality gives
\begin{align}\label{eq:DKW 2}
&\mathbb P\!\Big\{
\sup_y|\hat F_{\mu(\tau)}^\ell(y)-F_{\mu(\tau)}^\ell(y)|
\ge\sqrt{\frac{\ln(8T/\gamma)}{2\tilde m_\tau}}
\nonumber\\ 
&\hspace{4em}   \mid  \
\bar{\mathcal F}_\tau\vee\sigma(\tilde u_\tau)\Big\}
\le\frac{\gamma}{4T}.
\end{align} 
Taking expectations in \eqref{eq:DKW 2} and applying a
union bound over at most $2L_T\le2T$ pairs $(\tau,\ell)$ shows that the
corresponding DKW events hold simultaneously with probability at least
$1-\gamma/2$. 
Then, we have that
\begin{equation}\label{eq:DKW cvar mismatch 2}
|\bar\epsilon_{\mu(\tau)}|
\le\frac{4U}{\alpha}
\sqrt{\frac{\ln(8T/\gamma)}{2\tilde m_\tau}},
\qquad \tau=1,\ldots,L_T,
\end{equation} with probability at least $1-\gamma/2$. 

We next bound $R_1$. 
For each utilized packet, write $t=\mu(\tau)$. By the CVaR
gradient estimate and its estimation error, as in
\eqref{eq:estimated gradient 2} and
\eqref{eq:empirical error decomposition 2}, and using
$\|u_t\|=1$, we have
\begin{align*}
&\|\tilde{\bar g}_\tau\|
=\frac{d}{2\delta}\left|
C_t(x_t+\delta u_t)-C_t(x_t-\delta u_t)
+\bar\epsilon_t\right|\\
&\le\frac{d}{2\delta}\left|
C_t(x_t+\delta u_t)-C_t(x_t-\delta u_t)\right|
+\frac{d}{2\delta}|\bar\epsilon_t|\\
&\le dL_0+\frac{d}{2\delta}|\bar\epsilon_t|.
\end{align*} 
The second inequality follows from the $L_0$-Lipschitz continuity of $C_t$,
as in Lemma~\ref{lemma:smoothed cvar Lipschitz}. 
Substituting the uniform empirical-error bound
\eqref{eq:DKW cvar mismatch 2} into the last display yields
\begin{equation}\label{eq:two point gradient norm}
\|\tilde{\bar g}_\tau\|
\le dL_0+\frac{2dU}{\alpha\delta}
\sqrt{\frac{\ln(8T/\gamma)}{2\tilde m_\tau}}.
\end{equation} 
Following the proof of \eqref{eq:delay state mismatch}, we have
\begin{align*}
\|x_{\mu(\tau)}-\tilde x_\tau\| \le\eta\sum_{j=\tau-\tilde s_\tau}^{\tau-1}
\left(dL_0+\frac{2dU}{\alpha\delta}
\sqrt{\frac{\ln(8T/\gamma)}{2\tilde m_j}}\right).
\end{align*} 
Since $\tilde s_\tau\le2\bar d$, each virtual index occurs
in at most $2\bar d$ inner sums.
Hence,
\begin{equation}\label{eq:two point inverse sample sum}
\sum_{\tau=1}^{L_T}\frac{1}{\tilde m_\tau}
\le\sum_{t=1}^T\frac{1}{m_t}
\le\sum_{t=1}^T\frac{1}{\sqrt{m_t}}
\le cT^{1-a/2}.
\end{equation} 
The Cauchy--Schwarz inequality then implies
\begin{align*}
&\sum_{\tau=1}^{L_T}
\sum_{j=\tau-\tilde s_\tau}^{\tau-1}
\frac{1}{\sqrt{\tilde m_j}}\le\sqrt{D_T}
\left(2\bar d\sum_{\tau=1}^{L_T}\frac{1}{\tilde m_\tau}
\right)^{1/2}\\
&
\le\sqrt{2c\bar dD_T}\,T^{1/2-a/4}.
\end{align*}
Consequently,  
\begin{align}\label{eq:multi-update order 2}
R_1
&\le\eta dL_0^2D_T
+\frac{2\eta dUL_0}{\alpha\delta}
\sqrt{ c\bar dD_T\ln(8T/\gamma) }   \,T^{\frac{2-a}{4}}.
\end{align}

We now bound $R_2$. Define the ideal two-point estimator associated
with the $\tau$-th
utilized the packet by $\tilde g_\tau^{\rm tp}
 = \frac{d}{2\delta}
\big( C_{\mu(\tau)}(x_{\mu(\tau)}+\delta\tilde u_\tau)
 -C_{\mu(\tau)}(x_{\mu(\tau)}-\delta\tilde u_\tau)\big)
\tilde u_\tau.$ 
Since $x_{\mu(\tau)}$ is $\bar{\mathcal F}_\tau$-measurable and
$\tilde u_\tau$ is conditionally uniform and independent of
$\bar{\mathcal F}_\tau$, we have
\begin{equation}\label{eq:martingale difference 2}
\mathbb E[\tilde g_\tau^{\rm tp}\mid\bar{\mathcal F}_\tau]
=\nabla C_{\mu(\tau)}^\delta(x_{\mu(\tau)}).
\end{equation}
Let
$
\bar{\zeta}_\tau:=\tilde g_\tau^{\rm tp}
-\nabla C_{\mu(\tau)}^\delta(x_{\mu(\tau)}).
$
Here, $\bar\zeta_\tau$, $\tilde{\bar g}_\tau$, and
$\tilde{\bar\epsilon}_\tau$ are
$\bar{\mathcal F}_{\tau+1}$-measurable.
Then
\begin{equation}\label{eq:gradient decomposition 2}
\tilde{\bar g}_\tau
=\nabla C_{\mu(\tau)}^\delta(x_{\mu(\tau)})
+\bar{\zeta}_\tau
+\frac{d}{2\delta}\tilde{\bar\epsilon}_\tau\tilde u_\tau.
\end{equation} 
By the projected update and \eqref{eq:gt-x}, 
\begin{align}\label{eq:two-point projected sum}
\sum_{\tau=1}^{L_T}
\langle\tilde{\bar g}_\tau,\tilde x_\tau-x_{\mu(\tau)}^{\delta,\ast}\rangle
&\le R_{21}+\frac{\eta}{2}
\sum_{\tau=1}^{L_T}\|\tilde{\bar g}_\tau\|^2,
\end{align}  
where $R_{21}$ is the same telescoping term as in
\eqref{eq:R21}. 
Using $(p+q)^2\le2p^2+2q^2$, together with
\eqref{eq:two point gradient norm} and
\eqref{eq:two point inverse sample sum}, gives
\begin{align}\label{eq:two-point squared gradient sum}
\frac{\eta}{2}\sum_{\tau=1}^{L_T}
\|\tilde{\bar g}_\tau\|^2
 \le\eta d^2L_0^2T +\frac{2c\eta d^2U^2\ln(8T/\gamma)}{\alpha^2\delta^2}
 T^{1-a/2}.
\end{align}
Combining \eqref{eq:gradient decomposition 2} and
\eqref{eq:two-point projected sum}, using $\|\tilde x_\tau-x_{\mu(\tau)}^{\delta,\ast}\|\le D_x$, and
then applying \eqref{eq:two-point squared gradient sum}, we obtain
\begin{align}\label{eq:virtual order 2}
R_2
&\le R_{21}+\eta d^2L_0^2T
+\frac{2c\eta d^2U^2\ln(8T/\gamma)}{\alpha^2\delta^2}T^{1-a/2}\nonumber\\
&\quad+\bar R_{22}+\bar R_{23},
\end{align}
where $\bar R_{22}
 =\frac{dD_x}{2\delta}
\sum_{\tau=1}^{L_T}|\tilde{\bar\epsilon}_\tau|,$ and 
$\bar R_{23}
 =-\sum_{\tau=1}^{L_T}\langle\bar{\zeta}_\tau,\tilde x_\tau-x_{\mu(\tau)}^{\delta,\ast}\rangle.$  
Applying the sampling error summation used in the proof of
Theorem~\ref{theorem:one-point}, together with the two-batch error bound
\eqref{eq:DKW cvar mismatch 2}, gives 
\begin{align}\label{eq:estimation error order 2}
&\bar R_{22}
\le\frac{2dUD_x}{\alpha\delta}
\sqrt{\frac{\ln(8T/\gamma)}{2}}
\sum_{\tau=1}^{L_T}\frac{1}{\sqrt{\tilde m_\tau}}
\nonumber\\
&\le\frac{2cdUD_x}{\alpha\delta}
\sqrt{\frac{\ln(8T/\gamma)}{2}}T^{1-a/2}.
\end{align}

By \eqref{eq:martingale difference 2} and the same
Azuma--Hoeffding argument as in the proof of
Theorem~\ref{theorem:one-point}, using
$\|\tilde g_\tau^{\rm tp}\|\le dL_0$ and
$\|\nabla C_{\mu(\tau)}^\delta(x_{\mu(\tau)})\|\le L_0$, we obtain,
with probability at least $1-\gamma/2$, 
\begin{equation}\label{eq:direction martingale bound}
\bar R_{23}
\le(d+1)L_0D_x\sqrt{2L_T\ln\frac{2}{\gamma}}
\le(d+1)L_0D_x\sqrt{2T\ln\frac{2}{\gamma}}.
\end{equation}

Combining \eqref{eq:two-point virtual decomposition},
\eqref{eq:multi-update order 2}, \eqref{eq:virtual order 2},
 \eqref{eq:estimation error order 2}, and
\eqref{eq:direction martingale bound}, and applying a union bound, gives
\begin{align}\label{eq:two-point nonasymptotic bound}
&\mathrm{DR}_2(T)
\le\left(3+\frac{D_x}{r}\right)\delta L_0T
+2U\bar d+\eta dL_0^2D_T\nonumber\\
&\quad+\frac{2\eta dUL_0}{\alpha\delta}
\sqrt{ c\bar dD_T\ln(8T/\gamma)}  
 \,T^{\frac{2-a}{4}}+\eta d^2L_0^2T \nonumber\\
&\quad
+\!\frac{2c\eta d^2U^2\!\ln(8T/\gamma)}{\alpha^2\delta^2}
  T^{1\!-\!a/2} \!+\!\frac{2cdUD_x}{\alpha\delta}
\!\sqrt{\frac{\ln(8T/\gamma)}{2}}T^{1\!-\!a/2}\nonumber\\
&\quad+\frac{5D_x^2+4D_x\bar dP_T}{2\eta} +(d+1)L_0D_x\sqrt{2T\ln\frac{2}{\gamma}},
\end{align}
with probability at least $1-\gamma$.
Substituting the choices of $\delta$ and $\eta$ stated in the theorem
into \eqref{eq:two-point nonasymptotic bound}, and using
$D_T\le T+D_T$ and $L_T\le T$, we obtain Algorithm~\ref{alg: two-point}
achieves, with probability at least $1-\gamma$, 
$ \mathrm{DR}_2(T)
 =\widetilde{\mathcal O}\!\left(
T^{1-a/4}\sqrt{1+\bar dP_T}
 +(T+D_T)^{1/2}\sqrt{1+\bar dP_T+\bar d}
+\bar d\right).$ 
Moreover, when $P_T\ge1$, the regret bound simplifies to the result stated 
in \eqref{eq:two-point simplified path bound}.
\hfill$\blacksquare$

\begin{remark}\label{remark:common random samples}
Theorem~\ref{theorem:two-point} considers two mutually
independent sample batches in the estimator
\eqref{eq:estimated gradient 2}. If the two perturbed actions can
instead be evaluated using common random samples, namely,
$J_t(x_t+\delta u_t,\xi_t^i)$ and
$J_t(x_t-\delta u_t,\xi_t^i)$, then
Assumption~\ref{assumption:Lipschitz} gives
$\|\bar g_t\|\le dL_0$ pathwise. 
This leads to a refined regret bound of
\[
\widetilde{\mathcal O}\!\left(
T^{1-a/4}\sqrt{1+\bar dP_T}
+(T+D_T)^{1/2}\sqrt{1+\bar dP_T}+\bar d\right).
\]
Under $P_T\ge1$, it is the same as the simplified order
stated in Theorem~\ref{theorem:two-point}, but it removes the
corresponding finite-sample terms from the explicit bound.
\end{remark}

\begin{remark}\label{remark:one-two comparison}
Comparing Theorems~\ref{theorem:one-point}
and~\ref{theorem:two-point}, the two bounds have the same order at
$a=1$, while the two-point bound has a strictly smaller horizon
exponent for $1<a<2$. 
Since the two-point estimator queries two actions, 
it uses twice as many function evaluations compared to the one-point estimator.
However, this changes only a constant factor in the sampling requirement 
and therefore does not change the regret orders.
Furthermore, when $\bar d$ is independent of $T$, the two-point bound
approaches
$\widetilde{\mathcal O}(\sqrt{T(1+\bar dP_T)})$, which has the same
horizon and path-length order as the deterministic first-order delayed
OCO bound in \cite{wan2023non} as the sampling budget increases. This
comparison is only at the level of
regret order, since our setting involves stochastic CVaR objectives,
zeroth-order feedback, and finite-sample estimation errors.
\end{remark}

\section{Numerical Example}\label{sec:simulation}
\begin{figure}
    \centering
\includegraphics[width=0.96\linewidth]{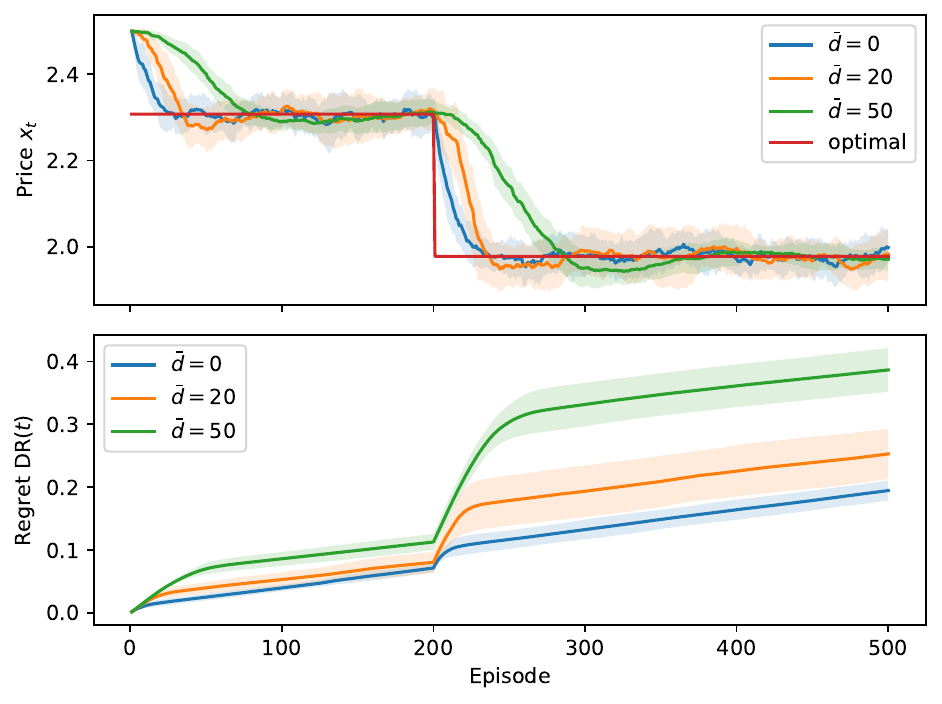}
    \caption{Effect of the maximum delay on the two-point method: 
    price trajectories and dynamic regret for $\bar d\in\{0,20,50\}$.}
    \label{fig:delay_compare}
\end{figure}

\begin{figure}
    \centering
    \includegraphics[width=0.96\linewidth]{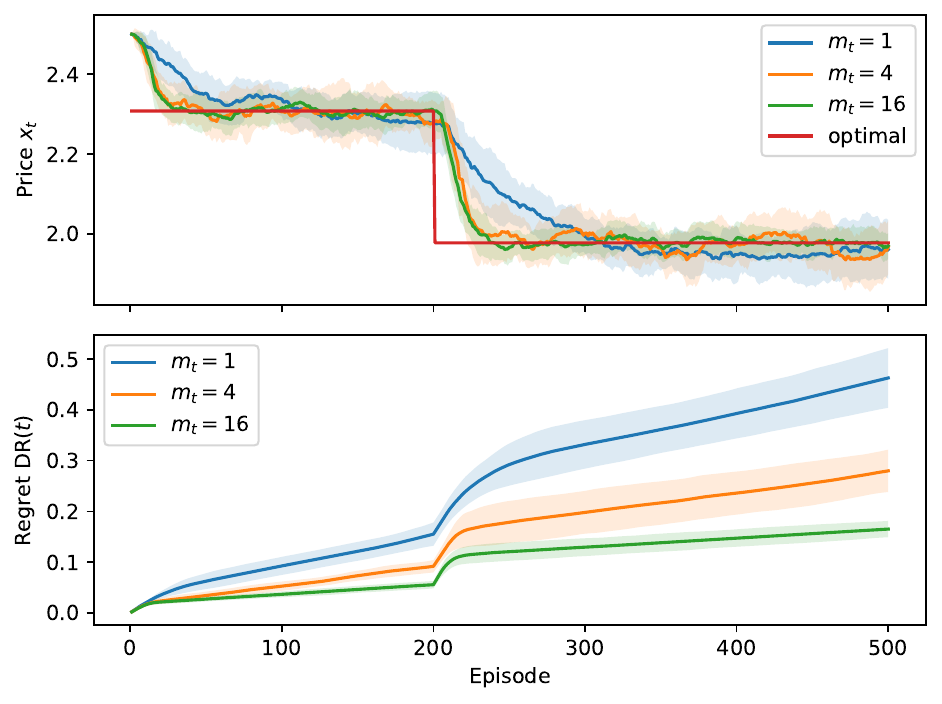}
    \caption{Effect of the batch size on the two-point method: 
    price trajectories and dynamic regret for the constant batch sizes $m_t\in\{1,4,16\}$.}
    \label{fig:sample_compare}
\end{figure}

In this section, we consider the dynamic pricing problem for parking lots \cite{ray2022decision}.  
Factors such as parking prices, availability, and locations generally influence driving decisions. This encourages us to dynamically adjust the parking price according to real-time demand. 
Denote $r_t \in [0,1]$ as curb occupancy rate. The occupancy rate is influenced by the price $x_t$ and environmental uncertainties $\xi_t$, which is given as $ r_t =\xi_t +A x_t$,
where $A = -0.15$ is the estimated price elasticity, determined by analyzing the real-world data \cite{ray2022decision}. 
Assume that the uncertainty $\xi_t$ follows the bounded uniform distribution $\mathcal U[0.9,1.1]$.  
Additionally, environmental factors such as weather and holidays may influence the desired occupancy rate $r_t^d$. 
To prevent either overcrowding or insufficient occupancy, our goal is to minimize the risk-averse objective function
\begin{equation*} 
C_t(x_t)=  \mathop{{\rm CVaR}_\alpha}\limits_{\xi_t \sim \mathcal U[0.9,1.1]} [J_t(x_t,\xi_t)],
\end{equation*}
where the risk level is set as $\alpha = 0.5$ and the loss function is given by 
\begin{equation*}
J_t(x_t,\xi_t)
     = \| \xi_t + A x_t -r_t^d
     \|^2+\frac{\nu}{2}\|x_t\|^2, 
\end{equation*}
where $\nu = 0.001$ is a regularization parameter. 
We use $\alpha=0.5$ in the baseline experiments as a moderate risk level that balances tail sensitivity and the stability of the empirical CVaR estimate at the baseline sampling budget. A more extreme risk level and rare-event uncertainty are considered separately below.
We set the initial price value as $x_0 = 2.5$ and restrict the potential prices to the range $[1.5,3]$.  
 \begin{figure}
    \centering
    \includegraphics[width=0.96\linewidth]{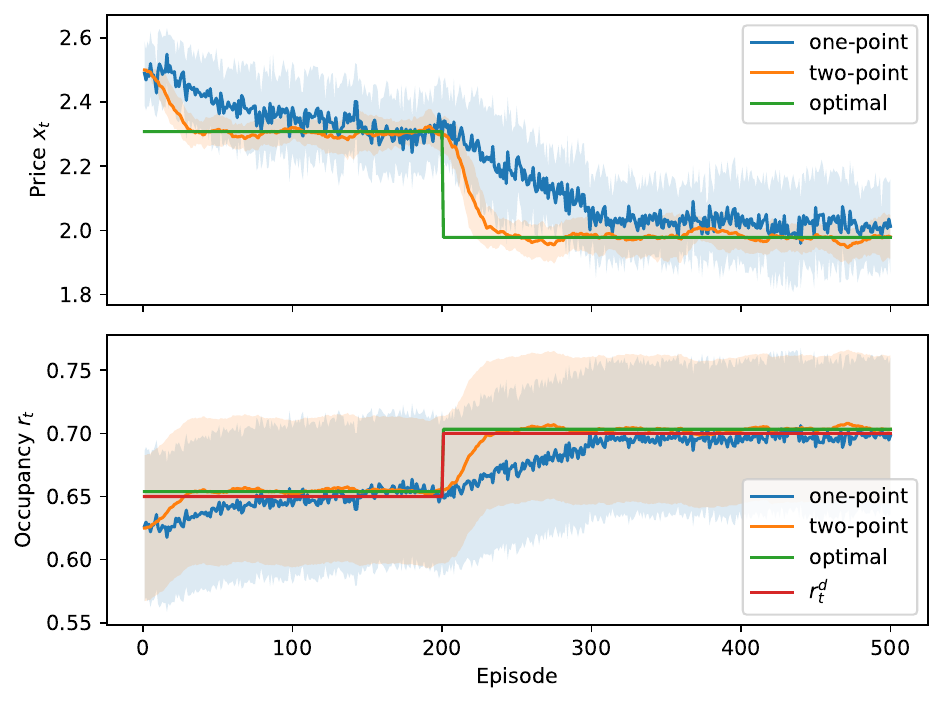}
    \caption{Price and occupancy trajectories of the one- and two-point methods under the step-changing target \eqref{eq:desired occupancy step}. 
    The optimal price and target occupancy are shown for reference.}
    \label{fig:price step}
\end{figure}
\begin{figure}
    \centering
    \includegraphics[width=0.96\linewidth]{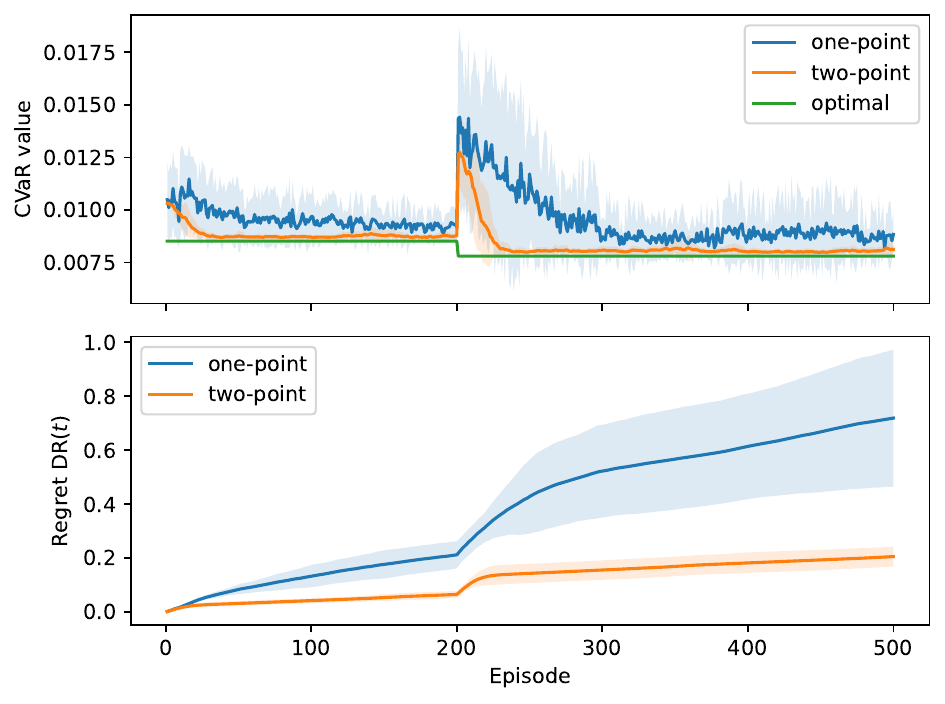}
    \caption{CVaR values and dynamic regret of the one- and two-point methods under the step-changing target 
    \eqref{eq:desired occupancy step}. The optimal CVaR value is shown for reference.}
    \label{fig:regret step} 
\end{figure}
\begin{figure}
    \centering
\includegraphics[width=0.96\linewidth]{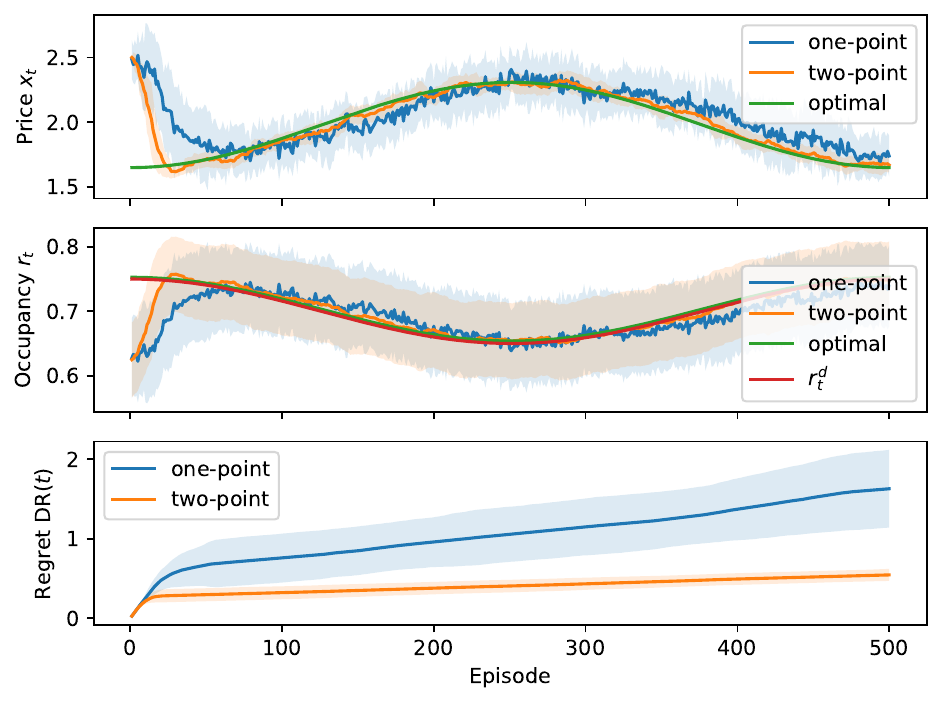}
    \caption{Price, occupancy, and dynamic regret of the one- and two-point methods 
    under the smoothly varying target $r_t^d=0.7+0.05\cos(t/80)$. 
    The optimal price and target occupancy are shown for reference.}
    \label{fig:price sin}
\end{figure}
The parking lot periodically monitors the occupancy rate and adjusts prices accordingly. 
Assume that the unit of time for an iteration is one week, which means the parking lot observes the occupancy rates and changes prices every week. However, due to data aggregation or computational overhead, these observations may be delayed.
Assume that the random delays  $d_t$ obey a uniform distribution with the range of $[0,10]$, with a maximum delay of $\bar{d} = 10$.  
For every direct comparison between the one- and two-point algorithms, 
both algorithms use $16$ stochastic function evaluations per round. 
For Figs.~\ref{fig:price step}--\ref{fig:sample_compare}, 
the algorithmic parameters are tuned separately for each displayed method and experimental setting. 
In the following figures,
shaded areas represent $\pm$ one standard deviation over $20$ runs.  
We consider the scenario where the desired occupancy changes over time due to variations in demand or usage patterns. 
Assume that the target occupancy rate is 
\begin{equation}\label{eq:desired occupancy step}
    r_t^d = \left\{\begin{array}{ll}
       0.65    &  {\text{if}}~t \in [1,200]  \\
       0.7  & {\text{if}}~ t \in (200,500].
    \end{array}\right.
\end{equation} 
Fig.~\ref{fig:price step} depicts the parking prices and corresponding occupancies generated by the one- and two-point risk-averse learning algorithms (Algorithms~\ref{alg: one-point} and \ref{alg: two-point}).
Fig.~\ref{fig:price step} shows that the prices and occupancies generated by Algorithm~\ref{alg: two-point} track desired values faster compared to 
Algorithm~\ref{alg: one-point}. 
Fig.~\ref{fig:regret step} depicts the CVaR values and the dynamic regret $\mathrm{DR}(t)$ 
of Algorithms~\ref{alg: one-point} and \ref{alg: two-point} under the changing desired occupancy, respectively. 
Fig.~\ref{fig:regret step} shows that the two-point algorithm outperforms the one-point algorithm by achieving a lower regret.
Moreover, it can be observed from  Figs.~\ref{fig:price step}--\ref{fig:regret step} that Algorithm \ref{alg: two-point} yields results with a lower variance compared to Algorithm \ref{alg: one-point}. This suggests that Algorithm \ref{alg: two-point}  produces more consistent and stable outcomes than Algorithm \ref{alg: one-point}. 

 Consider a time-varying target with a wider range 
 than the step change shown in Figs.~\ref{fig:price step}--\ref{fig:regret step}.
For instance, let the target occupancy be given by $r_t^d = 0.7 + 0.05\cos(t/80) $. Fig.~\ref{fig:price sin} depicts parking prices, the occupancies, and the dynamic regret $\mathrm{DR}(t)$ generated by Algorithms~\ref{alg: one-point} and \ref{alg: two-point}, respectively.  It can be observed that Algorithm~\ref{alg: two-point} achieves a faster convergence and smaller dynamic regret than Algorithm~\ref{alg: one-point}, even in the presence of more pronounced environmental fluctuations. 


Figs.~\ref{fig:delay_compare} and~\ref{fig:sample_compare} examine the effects of the maximum delay and batch size, 
respectively, on Algorithm~\ref{alg: two-point}. 
In Fig.~\ref{fig:delay_compare}, we set \(m_t=8\) and consider \(\bar d\in\{0,20,50\}\). 
As the delay increases, the price converges more slowly, 
while its variance and the dynamic regret increase. 
These effects are most pronounced at \(\bar d=50\). 
In Fig.~\ref{fig:sample_compare}, we fix \(\bar d=10\) and consider \(m_t\in\{1,4,16\}\). 
A larger batch size enables the price to track the optimal price more rapidly
 and reduces both the price variance and dynamic regret. 
 The algorithmic parameters are tuned separately for each delay or batch-size setting. 
 These observations are consistent with Theorem~\ref{theorem:two-point}, 
 which predicts improved performance with smaller delays or larger sampling budgets.

\begin{figure}
    \centering
\includegraphics[width=0.96\linewidth]{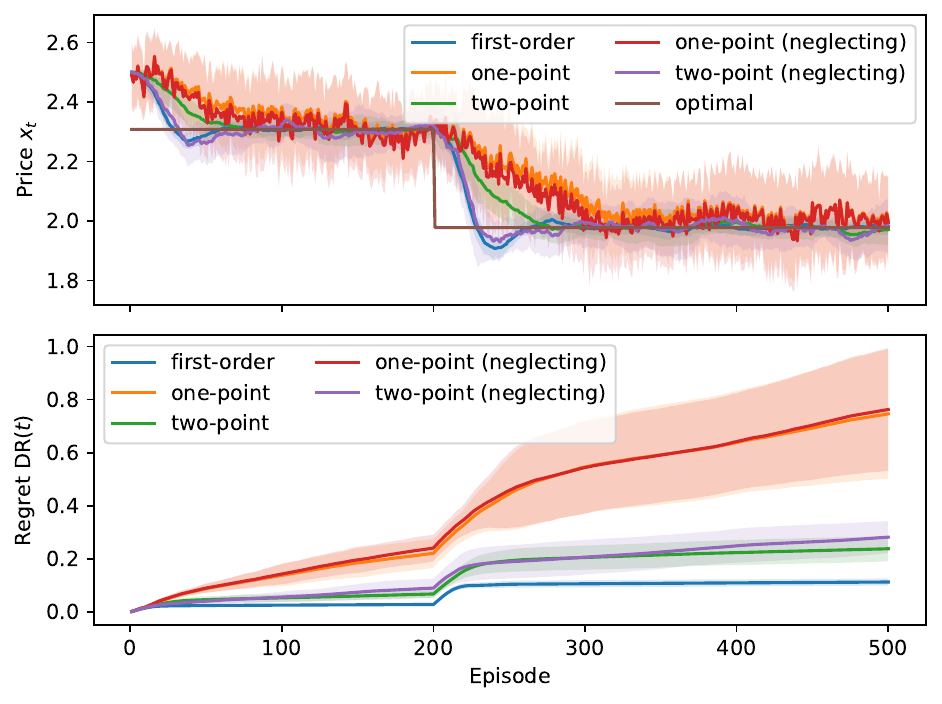}
    \caption{{}Price trajectories and dynamic regret under $\bar d=20$ for the delay-aware first-order (\cite{wang2025varying}),
     one-point (Alg.~\ref{alg: one-point}), and two-point methods (Alg.~\ref{alg: two-point}) and the delay-neglecting one- and two-point variants. 
     The optimal price is shown for reference.}
    \label{fig:compare_algs}
\end{figure} 
To provide a stronger reference, 
we adapt the sampled first-order empirical CVaR estimator in Algorithm~1 of \cite{wang2025varying} 
to delayed feedback. 
At each queried action, 
it estimates VaR from the sampled costs 
and averages the sample gradients in the empirical upper $\alpha$-tail to construct a CVaR gradient estimate. 
The first-order method uses $16$ sampled cost-gradient pairs per round. 
In Fig.~\ref{fig:compare_algs}, we set $\bar d=20$. 
Each method is tuned separately 
for its corresponding parameter-selection setting.
The delay-neglecting methods are tuned under the no-delay assumption. 
Figure~\ref{fig:compare_algs} shows that the first-order method performs best
 due to its access to gradient information. 
 Among the zeroth-order methods, two-point estimation yields lower dynamic regret 
 than one-point estimation under the same sampling budget, 
 while delay-aware tuning improves performance over the corresponding delay-neglecting implementations. 

\begin{figure}
    \centering
    \includegraphics[width=0.96\linewidth]{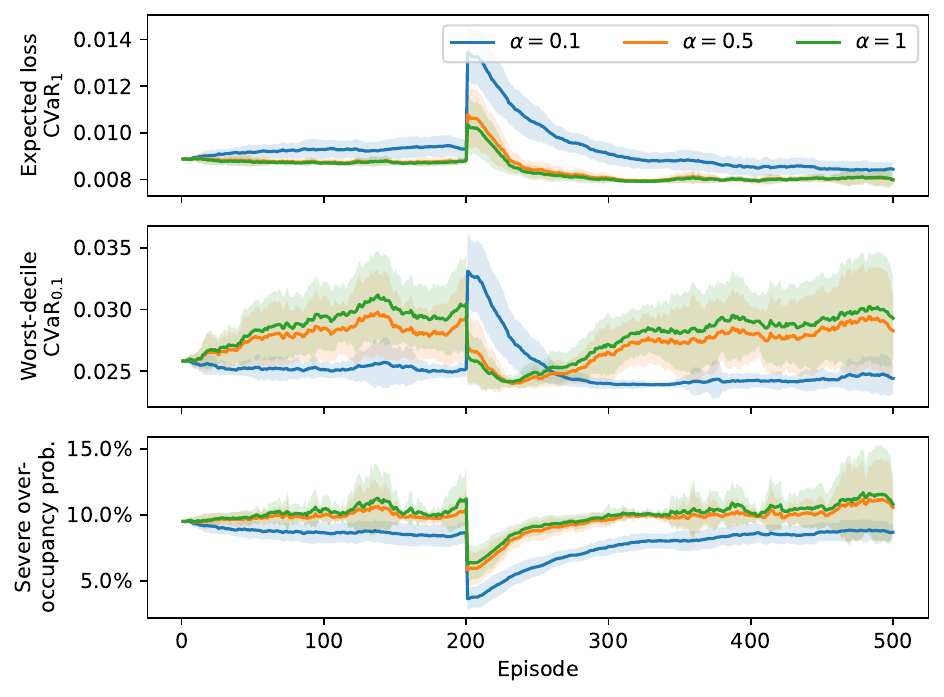}
    \caption{ Risk--performance tradeoff under bounded rare-event uncertainty: common expected loss $\mathrm{CVaR}_{1}$, worst-decile loss $\mathrm{CVaR}_{0.1}$, and severe over-occupancy probability $\Pr(r_t>r_t^d+0.1)$ for policies trained with $\alpha\in\{0.1,0.5,1\}$.} 
    \label{fig:risk_compare}
\end{figure} 
Figure~\ref{fig:risk_compare} provides a rare-event stress test. 
Specifically, $\xi_t\sim\mathcal U[0.9,1.1]$ with probability $0.9$, 
while $\xi_t\sim\mathcal U[1.12,1.22]$ with probability $0.1$, 
where the latter represents a rare high-demand event. 
This bounded mixture has a more pronounced upper-loss tail than the baseline distribution. 
We use the same sampling budget $m_t=8$, maximum delay $\bar d=10$, 
and delay realizations for all settings, 
and tune Algorithm~\ref{alg: two-point} separately for each risk level $\alpha\in\{0.1,0.5,1\}$. 
Here, a smaller $\alpha$ places more emphasis on extreme losses, 
while $\alpha=1$ reduces CVaR to the expected loss. 
For a fair comparison, 
all learned policies are evaluated using the same expected-loss measure $\mathrm{CVaR}_{1}$
 and the same tail risk measure $\mathrm{CVaR}_{0.1}$. 
 The bottom panel also reports the probability that occupancy exceeds its target by more than $0.1$, i.e., 
 $\Pr(r_t>r_t^d+0.1)$. 
 Figure~\ref{fig:risk_compare} shows that $\alpha=1$ attains the lowest expected loss 
 but the highest tail loss during the stationary periods, 
  whereas $\alpha=0.1$ exhibits the opposite behavior and $\alpha=0.5$ provides an intermediate tradeoff. 
In particular, compared with $\alpha=1$, the $\alpha=0.1$ policy reduces the 
stationary severe over-occupancy probability from approximately $10.5\%$ to $8.5\%$, 
a relative reduction of about $19\%$. 
Following the target change after $t=200$, 
the $\alpha=0.1$ policy incurs a larger transient tail loss, 
but after adapting it again achieves the lowest tail loss. 
Thus, stronger tail-risk protection provides a tangible reduction in overcrowding risk, 
at the cost of average performance and transient adaptation.

\section{Conclusion}\label{sec:conclusion}
This paper investigated risk-averse learning with delayed feedback, 
where the delays are random yet bounded.  
We developed risk-averse learning algorithms based on one- and
two-point batch zeroth-order estimators, 
which draw a batch of stochastic function
evaluations to form the empirical CVaR and its gradient.  
We sorted the arriving feedback according to their utilization order and calculated the cumulative delay. 
 The dynamic regret bounds of these algorithms were then analyzed in terms of the cumulative delay, 
 the total number of samples, and the path length of comparators. 
 When the maximum delay is independent of the horizon and
$P_T=\mathcal O(1)$, both algorithms achieve sublinear regret under the
stated sampling requirement. As the sample size increases, the
two-point method can attain a smaller regret order than the one-point
method. With sufficiently large sample sizes, the two-point bound approaches the
deterministic first-order delayed OCO benchmark in \cite{wan2023non}.
Future work will explore meta-algorithmic frameworks for adaptive
risk-averse learning when the delay bound and comparator variation are unknown.

\section{Appendix}\label{sec:appendix}

\noindent\textit{Proof of Lemma~\ref{lemma:delay}.}  
Every packet generated at time $t\le T-\bar d$ arrives by
the end of the horizon. Hence, $L_T\ge T-\bar d$.
Under the condition $T\ge3\bar d+1$ in the second claim,
we further have $L_T\ge T-\bar d\ge2\bar d+1$, so the index ranges used
in its proof are valid.
The first claim follows from \cite{li2019bandit}, but we provide its proof here for completeness. 
Denote $ L_{\mu(\tau)-1} = m $, which means that the learner receives $m$ batches feedback before $t_1=\mu(\tau)$.
Due to the delay, we have $0  \le m \le \mu(\tau)-1$. 
The learner receives the $\tau$-th batch feedback at slot $t_2 = \mu(\tau)+d_{\mu(\tau)} \ge t_1$.
 In other words, the learner receives at least $m$ batches feedback before $t_2$. 
 Then, we have $\tau \ge m+1$, and further that $\tilde{s}_\tau \ge m+1-1-m = 0$. 
 Additionally, before the slot $\mu(\tau)$, the learner receives $L_{\mu(\tau)-1}$ batches feeback. 
 Note that $L_{\mu(\tau)-1} \ge \mu(\tau)-\bar{d} -1$, 
 we have $    \tilde{s}_\tau = \tau-1-L_{\mu(\tau)-1} \le \tau -1 - \mu(\tau) +1+\bar{d} \le 2\bar{d}. $ 

We next prove the second claim. According to the definition of $\tilde{s}_\tau$, we have 
\begin{align}\label{eq:sum s tau}
  &\sum_{\tau=1}^{L_T} \tilde{s}_\tau = \sum_{\tau=1}^{L_T} \big(\tau - 1 - L_{\mu(\tau)-1}\big) \nonumber \\ 
    & = \sum_{t=1}^{L_T} (t-1) - \sum_{\tau=1}^{L_T}L_{\mu(\tau)-1} \nonumber \\ 
    & = \sum_{t=1}^{L_T} (t-1) - \sum_{t=1}^{L_T-\bar{d}}L_{t-1} -  \sum_{\tau\in \mathcal{S}} L_{\mu(\tau)-1},
\end{align}
with $\mathcal{S} = \{ \tau| 1\le \tau \le L_T, \mu(\tau)>L_T-\bar{d} \}$. 
For the first and the second terms of \eqref{eq:sum s tau}, we have 
\begin{align}\label{eq: sum s tau 1}
& \sum_{t=1}^{L_T} (t-1) - \sum_{t=1}^{L_T-\bar{d}}L_{t-1} \nonumber \\ 
&= \sum_{t=1}^{L_T-2\bar{d}-1} t - \sum_{t=1}^{L_T-\bar{d}-1}L_t + \sum_{t=L_T-2\bar{d}}^{L_T-1} t \nonumber \\ 
&\le  \sum_{t=1}^{L_T-2\bar{d}-1}d_t  + \sum_{t=L_T-2\bar{d}}^{L_T-1} t \nonumber \\ 
&= \sum_{t=1}^{L_T-2\bar{d}-1} d_t +   \bar{d}(2L_T-2\bar{d}-1),  
\end{align}
where the first equality follows from $L_0 = 0$. We next explain the inequality. In the absence of delay, the learner receives $t$ batches of feedback by the end of time slot $t$. Then, $\sum_{t=1}^{{L_T-\bar{d}-1}} t$ can be interpreted as each packet is accounted once at each time slot since its arrival over the horizon $L_T-\bar{d}-1$. Accordingly, $\sum_{t=1}^{{L_T-\bar{d}-1}} L_t$ means that each packet is accounted once at each time since its arrival in the presence of delay.  
Additionally, due to the delay, each $J_t(\cdot)$ is undercounted $d_t$ times by time slot $t+\bar{d}$. 
By Assumption~\ref{ass:max delay}, the losses $J_t(\cdot)$, for $t \in \{1,\dots, L_T-2\bar{d}-1\}$, arrive before the end of time $ L_T-\bar{d}-1$. 
Thus, we have $\sum_{t=1}^{L_T-2\bar{d}-1} t - \sum_{t=1}^{L_T-\bar{d}-1}L_t \le \sum_{t=1}^{L_T-2\bar{d}-1} d_t $. 
For the last term of 
\eqref{eq:sum s tau}, we have
\begin{align}\label{eq:sum s tau 2}
  &  \sum_{\tau\in \mathcal{S}} L_{\mu(\tau)-1} \ge \sum_{\tau\in\mathcal{S}} \mu(\tau)-\bar{d}-1 \nonumber \\ 
    &\ge \sum_{L_T-2\bar{d}}^{L_T-\bar{d}-1} t = \frac{\bar{d}(2L_T-3\bar{d}-1)}{2},
\end{align}
where the first inequality follows from $L_{t} \ge t-\bar{d}$ and the second inequality follows from the fact the set $\mathcal{S}$  has $\bar{d}$ components and $\mu(\tau) \ge L_T-\bar{d}+1$ for $\tau \in \mathcal{S}$, as in \eqref{eq:sum s tau}. 
Substituting \eqref{eq: sum s tau 1}--\eqref{eq:sum s tau 2} into \eqref{eq:sum s tau}, we obtain the second claim. 
$\hfill$ $\blacksquare$

The algorithm's decisions are compared against the per-iteration optimum. Given that delays cause feedback to arrive out of order, it is necessary to consider the associated shifts in the comparator. We provide the following lemma to address this issue, and it is used in the proof of both Theorems~\ref{theorem:one-point} and~\ref{theorem:two-point}.
 
\begin{lemma}\label{lemma:comparator}
Under Assumption~\ref{ass:max delay}, the reordered comparator sequence
satisfies
\begin{equation*}
\sum_{\tau=2}^{L_T}
\bigl\|
x_{\mu(\tau)}^\ast-x_{\mu(\tau-1)}^\ast
\bigr\|
\le
2\bar d P_T .
\end{equation*}
\end{lemma}

\noindent\textit{Proof of Lemma~\ref{lemma:comparator}.}
First, we establish the displacement between the virtual index
$\tau$ and the generation time $\mu(\tau)$. Let $\mu(\tau)=t$.
The feedback generated at time $t$ arrives at time $t+d_t$.
Since at most $t+d_t$ feedback batches have been generated by that
time, its utilization rank satisfies
$
\tau\le t+d_t\le t+\bar d,
$
and hence
$
\mu(\tau)=t\ge\tau-\bar d.
$

For the reverse inequality, every feedback batch generated at time
$k\le t-\bar d$ satisfies
$
k+d_k\le k+\bar d\le t\le t+d_t.
$
Such a batch is therefore utilized before the batch generated at time
$t$. If two batches arrive simultaneously, their utilization order is
their generation order. Consequently,
$
\tau-1\ge t-\bar d,
$
whenever $t>\bar d$; for $t\le\bar d$, the resulting upper bound is
immediate. Thus,
\begin{equation}\label{eq:rank displacement}
\tau-\bar d
\le
\mu(\tau)
\le
\tau+\bar d-1.
\end{equation}
For every $\tau=2,\ldots,L_T$, the triangle inequality gives
\begin{align*}
\bigl\|
x_{\mu(\tau)}^\ast-x_{\mu(\tau-1)}^\ast
\bigr\| \le
\sum_{k=\min\{\mu(\tau-1),\mu(\tau)\}+1}
     ^{\max\{\mu(\tau-1),\mu(\tau)\}}
\bigl\|x_k^\ast-x_{k-1}^\ast\bigr\|.
\end{align*}

Consider a fixed increment
$\|x_k^\ast-x_{k-1}^\ast\|$.
If it is crossed in the forward direction, then
$
\mu(\tau-1)<k\le\mu(\tau).
$
Using \eqref{eq:rank displacement}, this can occur only when
$
k-\bar d+1\le\tau\le k+\bar d.
$
If it is crossed in the reverse direction, then
$
\mu(\tau)<k\le\mu(\tau-1),
$
which can occur only when
$
k-\bar d+2\le\tau\le k+\bar d-1.
$
Therefore, each increment
$\|x_k^\ast-x_{k-1}^\ast\|$ appears in at most $2\bar d$ of the
preceding sums. It follows that
\begin{align*}
\sum_{\tau=2}^{L_T}
\bigl\|
x_{\mu(\tau)}^\ast-x_{\mu(\tau-1)}^\ast
\bigr\|
 \le
2\bar d
\sum_{k=2}^{T}
\bigl\|x_k^\ast-x_{k-1}^\ast\bigr\| =
2\bar d P_T.
\end{align*}
\hfill$\blacksquare$

\bibliographystyle{ieeetr}
\bibliography{references}

\end{document}